\documentclass[11pt]{article}
\pdfoutput=1
\usepackage{amsmath,amsfonts,amssymb,cases, textcomp}
\usepackage[T1]{fontenc}
\usepackage[utf8]{inputenc}

\usepackage[dvipsnames]{xcolor}

\definecolor{my_colour2}{rgb}{0,0,0} 	 

\usepackage{array}
\newcommand{\PreserveBackslash}[1]{\let\temp=\\#1\let\\=\temp}
\newcolumntype{C}[1]{>{\PreserveBackslash\centering}p{#1}}
\newcolumntype{R}[1]{>{\PreserveBackslash\raggedleft}p{#1}}
\newcolumntype{L}[1]{>{\PreserveBackslash\raggedright}p{#1}}
\usepackage{textcomp}
\usepackage{stfloats}
\usepackage{url}
\usepackage{verbatim}
\usepackage{graphicx}
\usepackage{balance}

\usepackage{footnote}
\usepackage{float}

\usepackage[linesnumbered,ruled,vlined]{algorithm2e}
\DontPrintSemicolon

\SetKwComment{Comment}{\color{green!50!black}// }{}
\renewcommand{\CommentSty}[1]{\textnormal{\ttfamily\color{green!50!black}#1}\unskip}

\SetKwProg{Function}{function}{}{}

\usepackage{physics}

\usepackage{booktabs}
\usepackage{colortbl}
\usepackage{multirow}

\usepackage{setspace}
\usepackage{algpseudocode}
\usepackage{enumitem}

\usepackage{tikz}
\usepackage{tikzscale}
\usetikzlibrary{positioning}
\usetikzlibrary{fit}
\usetikzlibrary{patterns}
\usetikzlibrary{patterns.meta}
\usetikzlibrary{arrows.meta}
\usetikzlibrary{shapes}
\usetikzlibrary{calc}

\usepackage{pgfplots}
\pgfplotsset{compat=1.15}

\usepackage{amsthm}

\newtheorem{theorem}{Theorem}
\newtheorem{proposition}[theorem]{Proposition}
\newtheorem{lemma}[theorem]{Lemma}

\usepackage{caption}
\usepackage{subcaption}
\usepackage{upgreek,bm} 

\newcommand{\vv}[1]{\textbf{#1}}
\newcommand{\vvv}[2]{\textbf{#1}^{(#2)}}
\newcommand{\vvh}[2]{\hat{\textbf{#1}}^{(#2)}}
\newcommand{\erre}[1]{\mathbb{R}^{#1}}
\newcommand{\boldtheta}{\bm{\uptheta}}
\newcommand{\boldphi}{\bm{\upvarphi}}

\newcommand{\loss}{L}
\newcommand{\actf}{\mathcal{A}}
\newcommand{\mygradient}[2]{\dv{#1}{#2}}

\newcommand{\dvinline}[2]{ {\dd{#1}} \big{/} {\dd{#2}} }
\newcommand{\pdvinline}[2]{ {\partial #1} \big{/} {\partial #2} }

\usepackage[authoryear]{natbib}
\bibpunct{(}{)}{;}{a}{,}{,}

\newif\iftitoletti
\newcommand{\titoletto}[1]{\iftitoletti \noindent $\rhd$ \textbf{#1} \\ \fi}
\titolettifalse

\usepackage[dvipsnames]{xcolor}
\definecolor{mydarkgreen}{rgb}{0.0, 0.5, 0.0}


\begin{document}

\newlength\tindent
\setlength{\tindent}{\parindent}
\renewcommand{\indent}{\hspace*{\tindent}}
\setlength{\parindent}{10pt}

\setstretch{1.25}

\begin{savenotes}
\title{
\bf{ 
RNN($p$) for
Power Consumption Forecasting
}}

\author{
Roberto Baviera$^{1}$
\& Pietro Manzoni$^{2}$
}

\maketitle

\vspace*{0.08truein}
\begin{tabular}{ll}
 &  $^1$ Politecnico di Milano, Department of Mathematics, Italy \\
 &  $^2$ University of Edinburgh, Business School, United Kingdom \\
\end{tabular}
\end{savenotes}

\vspace*{0.20truein}

\begin{abstract}

\noindent
An elementary Recurrent Neural Network that operates on $p$ time lags, called an RNN($p$), is the natural generalisation of a linear autoregressive model ARX($p$).
It is a powerful forecasting tool for variables displaying inherent seasonal patterns across multiple time scales, as is often observed in energy, economic, and financial time series.

\noindent
The architecture of RNN($p$) models, characterised by structured feedbacks across time lags, enables the design of efficient training strategies.
%
We conduct a comparative study of learning algorithms for these models, providing
a rigorous analysis of their computational complexity and training performance.

\noindent
We present two applications of RNN($p$) models in power consumption forecasting, a key domain within the energy sector where accurate forecasts inform both operational and financial decisions. 
Experimental results show that RNN($p$) models achieve excellent forecasting accuracy while maintaining a high degree of interpretability. These features make them well-suited for decision-making in energy markets and other fintech applications where reliable predictions play a significant economic role.

\noindent
\end{abstract}

\vspace*{0.11truein}

\iftrue
	\vspace{0.8cm}
	\noindent
	{\small
	{\bf Declarations}:
	The authors declare that they have no financial or non-financial interests related to this work and 
	received no external support for its development.
	}
	\vspace{1cm}
\fi

\begin{flushleft}
{\bf Address for correspondence:}\\[0.3cm]

{\bf Pietro Manzoni}\\
University of Edinburgh\\ 
Business School \\
29 Buccleuch Place, EH8 9JS \\ 
Edinburgh, UK \\
pmanzoni@ed.ac.uk\\
\end{flushleft}

\newpage

\setstretch{1.30}


\section{Introduction} \label{sec:introduction}

\titoletto{RNNs: what they are}
Recurrent Neural Networks (RNNs) are a powerful class of Artificial Neural Networks specifically designed to 
learn patterns in sequences of data. Unlike traditional Feedforward Neural Networks (FNNs), which treat all data points as independent, RNNs are structured
to capture 
sequential dependencies and contextual information.
For this reason, they are particularly well-suited for tasks involving time series data \citep[see, e.g.,][]{lim2021time}.

\titoletto{RNN for Time series Forecasting: Cosa viene fatto adesso}
This paper explores the application of RNNs to time series forecasting.
The ability to properly identify
meaningful time dependencies
is crucial for making accurate predictions.
In the literature on RNN-based time series forecasting, the 
state-of-the-art models are Long Short-Term Memory (LSTM) and Gated Recurrent Unit (GRU) \citep[see, e.g.,][]{hewamalage2021}.
These networks are designed to learn even non-trivial time dependencies 
by utilising
gating mechanisms to regulate the flow of information and maintain a persistent memory \citep[see, e.g.,][]{goodfellow2016}.

\titoletto{Drawbacks of LSTM and GRU}
Despite their advanced capabilities, the architectures of LSTM and GRU are notably complex, as they involve multiple gates and state variables. While these features contribute to network effectiveness, they also increase computational complexity and pose significant challenges in practical applications, particularly in terms of training and interpretability\footnote{
We recall that,
as defined by the ISO/IEC standards for AI-based systems,
the notion of \textit{model interpretability} refers to ``the level of understanding how the underlying (AI) technology works'' \citep[cf.][3.1.42]{ISO29119}.} \citep[see, e.g.,][]{aggrawal2018}.

\medskip
\titoletto{Cosa Proponiamo in Generale}
Some time series, especially in the energy sector, are characterised by an inherent seasonal structure, as a result of both natural factors (e.g., temperature and solar radiation on daily and yearly cycles) and human activities (e.g., daily and weekly cycles tied to working hours). By leveraging these intrinsic dependencies, our goal is to design an RNN with an expert-informed architecture that incorporates temporal feedbacks 
aligned with the known seasonalities.
We aim to show that this
approach can significantly enhance:
i) the interpretability of the model;
ii) in some cases, even its predictive performance.

\titoletto{Cosa on top di Existing (Contribution I)}
In this paper, we introduce \textit{RNN(p) models}, a family of
\textit{multi-lag} RNNs, i.e.\ of RNNs that integrate multiple feedbacks from preceding time steps \citep[see, e.g.,][]{dejesus2007}.\footnote{Multi-lag RNNs generalise the simpler, widely used
\textit{single-lag} RNNs,
which incorporate only one feedback loop and
have proven effective in capturing very short-term dependencies
\citep[see, e.g.,][]{goodfellow2016}.}
RNN($p$) models represent the simplest nonlinear extension of the linear autoregressive models with exogenous inputs, known as ARX($p$) in the literature \citep[see, e.g.,][]{box2015}.
They feature an elementary architecture, consisting of a single hidden layer and multiple Jordan feedbacks -- i.e.\ feedbacks from the output layer to the hidden layer \citep[cf.][]{jordan1986}.
\titoletto{Commenting on the Figure}
Figure \ref{fig:fig1} illustrates a schematic diagram of the model architecture. On the left, we show the structure of an ARX($1$) alongside that of an RNN($1$); as we observe, the latter incorporates an additional hidden layer 
$\vvv{h}{t}$ which enables the modelling of nonlinearities. On the right, we show the general multi-lag case, where feedback connections from the previous $p$ time steps are included. 

\begin{figure}[!t]
	\centering
	\begin{subfigure}{.5\textwidth}
  		\centering
		\begin{tikzpicture}[->,draw=black!75, scale=1, transform shape]

    \filldraw[fill=green!30, draw = mydarkgreen,  rounded corners=8, line width=0.25mm] (0.0,0.90) circle(0.5);
    \node[] at (0.1, 0.95) {$\textbf{x}^{(t)}$};

    \filldraw[fill=red!30, draw = red,  rounded corners=8, line width=0.25mm] (0.0,2.8) circle(0.5);
    \node[] at (0.1,2.77) {$\hat{\textbf{y}}^{(t)}$};
    \node[] at (1.65,2.77) {$\hat{\textbf{y}}^{(t-1)}$};

    \draw [line width=0.35mm] (0.0, 1.45) -- (0.0, 2.25);

	\draw[-, smooth, dashed, line width=0.35mm] 
		(0.00, 3.35)	to [out=90, in=-90] 	(0.00, 3.50) -- 
		(0.00, 3.50)	to [out=90, in=-170] 	(0.35, 4.00) -- 
		(0.35, 4.00) 	to [out=10, in=-190] 	(0.60, 4.00) -- 
		(0.60, 4.00) 	to [out=-10, in=90] 	(1.00, 2.75);
		
	\draw[-, smooth, line width=0.35mm]
		(1.00, 2.75) 	to [out=-90, in=10] 	(0.50, 1.70) --
		(0.50, 1.70) 	to [out=-170, in=-10] 	(0.35, 1.70) --
		(0.35, 1.70) 	to [out=-190, in=-90] 	(0.00, 2.109);


    \filldraw[fill=green!30, draw = mydarkgreen,  rounded corners=8, line width=0.25mm] (3.0,0.90) circle(0.5);
    \node[] at (3.10, 0.95) {$\textbf{x}^{(t)}$};

    \filldraw[fill=cyan!30, draw = cyan, rounded corners=8, line width=0.25mm] (2.55, 2.95)  rectangle ++(0.9,-0.9){};
    \node[] at (3.10, 2.55) {$\textbf{h}^{(t)}$};

    \filldraw[fill=red!30, draw = red,  rounded corners=8, line width=0.25mm] (3.0,4.02) circle(0.5);
    \node[] at (3.10,4.02) {$\hat{\textbf{y}}^{(t)}$};
    \node[] at (4.7,3.30) {$\hat{\textbf{y}}^{(t-1)}$};

    \draw [line width=0.35mm] (3.0, 1.45) -- (3.0, 2.02);
    \draw [line width=0.35mm] (3.0, 3.00) -- (3.0, 3.47);

	\draw[-, smooth,  dashed, line width=0.35mm] 
		(3.00, 4.58)	to [out=90, in=-170] 	(3.30, 4.90) -- 
		(3.30, 4.90) 	to [out=10, in=-185] 	(3.60, 4.90) -- 
		(3.60, 4.90) 	to [out=-05, in=90] 	(4.05, 3.70) --
		(4.05, 3.70) 	to [out=-90, in=90] 	(4.05, 3.30);

	\draw[-, smooth, line width=0.35mm]
		(4.05, 3.30) 	to [out=-90, in=90] 	(4.05, 2.60) --
		(4.05, 2.60) 	to [out=-90, in=10] 	(3.50, 1.60) --
		(3.50, 1.60) 	to [out=-170, in=-10] 	(3.30, 1.60) --
		(3.30, 1.60) 	to [out=-190, in=-90] 	(3.00, 1.90);

    \node[] at (0.3, 5.6) {\textbf{ARX($1$)}};
    \node[] at (3.3, 5.6) {\textbf{RNN($1$)}};

    \node[] at (0.3, 0.2) {\phantom{---}};

\end{tikzpicture}		
  		\caption{\centering{\small{
  		Autoregressive models of order $1$ in the linear \newline 
  		(ARX, \textit{left}) and nonlinear (RNN, \textit{right}) cases.}}}
  		\label{fig:sub1}
	\end{subfigure}%
	\begin{subfigure}{.5\textwidth}
  		\centering
\begin{tikzpicture}[->,draw=black!75]

    \filldraw[fill=green!30, draw = mydarkgreen, line width=0.25mm] (-1,0.9) circle(0.5);
    \node[] at (-0.9, 0.95) {$\textbf{x}^{(t)}$};

    \filldraw[fill=red!30, draw = red, line width=0.25mm] (-1,2.8) circle(0.5);
    \node[] at (-0.9, 2.77) {$\hat{\textbf{y}}^{(t)}$};

    \draw [line width=0.35mm] (-1, 1.45) -- (-1, 2.25);

    \draw[-, dashed, smooth, line width=0.35mm] 
        (-1.00, 3.35) 	to [out=90, in=-90] 	(-1.00, 3.50) --
        (-1.00, 3.50) 	to [out=90, in=-170] 	(-0.50, 4.00) --
        (-0.50, 4.00) 	to [out=10, in=-190] 	(-0.35, 4.00) --
        (-0.35, 4.00) 	to [out=-10, in=90] 	( 0.15, 2.80);

    \draw[-, smooth, line width=0.35mm] 
   		( 0.13, 3.40) 	to [out=-87, in=90] 	( 0.15, 2.80) --
        ( 0.15, 2.80) 	to [out=-90, in=10] 	(-0.35, 1.70) --
        (-0.35, 1.70) 	to [out=-170, in=-5] 	(-0.50, 1.70) --
        (-0.50, 1.70) 	to [out=-185, in=-90] 	(-1.00, 2.10);

    \draw[-, dashed, smooth, line width=0.35mm] 
        (-1.00, 3.35) 	to [out=90, in=-90] 	(-1.00, 3.50) -- 
        (-1.00, 3.50) 	to [out=90, in=-170] 	(-0.30, 4.10) --
        (-0.30, 4.10) 	to [out=10, in=-190] 	( 0.35, 4.10) -- 
        ( 0.35, 4.10) 	to [out=-10, in=90] 	( 0.90, 2.80);

    \draw[-, smooth, line width=0.35mm]
        ( 0.90, 2.80) 	to [out=-90, in=10] 	( 0.35, 1.60) --
        ( 0.35, 1.60) 	to [out=190, in=-10] 	(-0.30, 1.60) --
        (-0.30, 1.60) 	to [out=170, in=-90] 	(-1.00, 2.10);

    \draw[-, dashed, line width=0.35mm] 
        (-1.00, 3.35) 	to [out=90, in=-90] 	(-1.00, 3.50) -- 
        (-1.00, 3.50) 	to [out=90, in=-170] 	(-0.30, 4.10) --
        (-0.30, 4.10) 	to [out=10, in=-190] 	( 0.90, 4.10) -- 
        ( 0.90, 4.10) 	to [out=-10, in=90] 	( 1.50, 2.80) --
        ( 1.50, 2.80) 	to [out=-90, in=80] 	( 1.40, 2.10);

    \draw[-, smooth, line width=0.35mm] 
        ( 1.40, 2.10) 	to [out=-100, in=10] 	( 0.90, 1.60) --
        ( 0.90, 1.60) 	to [out=190, in=-10] 	(-0.30, 1.60) --
        (-0.30, 1.60) 	to [out=170, in=-90] 	(-1.00, 2.10);

    \node[fill=white,inner sep=1pt] at (0.10, 3.50) {$\hat{\textbf{y}}^{(t-1)}$}; 
    \node[fill=white,inner sep=1pt] at (0.85, 2.90) {$\hat{\textbf{y}}^{(t-2)}$}; 
    \node[fill=white,inner sep=1pt] at (1.55, 2.20) {$\hat{\textbf{y}}^{(t-p)}$};


    \filldraw[fill=green!30, draw = mydarkgreen,  rounded corners=8, line width=0.25mm] (3.0,0.90) circle(0.5);
    \node[] at (3.10, 0.95) {$\textbf{x}^{(t)}$};

    \filldraw[fill=cyan!30, draw = cyan, rounded corners=8, line width=0.25mm] (2.55, 2.95)  rectangle ++(0.9,-0.9){};
    \node[] at (3.10, 2.55) {$\textbf{h}^{(t)}$};

    \filldraw[fill=red!30, draw = red,  rounded corners=8, line width=0.25mm] (3.0,4.02) circle(0.5);
    \node[] at (3.10,4.02) {$\hat{\textbf{y}}^{(t)}$};

    \draw [line width=0.35mm] (3.0, 1.45) -- (3.0, 2.02);
    \draw [line width=0.35mm] (3.0, 3.00) -- (3.0, 3.47);




	\draw[-, dashed, line width=0.35mm] 
		(3.00, 4.58)	to [out=90, in=-170] 	(3.30, 4.90) -- 
		(3.30, 4.90) 	to [out=10, in=-185] 	(3.80, 4.90) -- 
		(3.80, 4.90) 	to [out=-05, in=90] 	(4.20, 3.70) --
		(4.20, 3.70) 	to [out=-90, in=90] 	(4.20, 2.60) --
		(4.20, 2.60) 	to [out=-90, in=10] 	(3.70, 1.60) --
		(3.70, 1.60) 	to [out=-170, in=-10] 	(3.30, 1.60) --
		(3.30, 1.60) 	to [out=-190, in=-90] 	(3.00, 1.90);
    \draw[-, smooth, line width=0.35mm]
		(4.20, 3.70) 	to [out=-90, in=90] 	(4.20, 2.60) --
		(4.20, 2.60) 	to [out=-90, in=10] 	(3.70, 1.60) --
		(3.70, 1.60) 	to [out=-170, in=-10] 	(3.30, 1.60) --
		(3.30, 1.60) 	to [out=-190, in=-90] 	(3.00, 1.90);

	\draw[-, dashed, line width=0.35mm] 
		(3.00, 4.58)	to [out=90, in=-175] 	(3.30, 4.90) -- 
		(3.30, 4.90) 	to [out=05, in=-185] 	(4.40, 4.90) -- 
		(4.40, 4.90) 	to [out=-05, in=90] 	(4.85, 3.50);
	\draw[-, smooth, line width=0.35mm] 
		(4.85, 3.50) 	to [out=-90, in=90] 	(4.85, 2.60) --
		(4.85, 2.60) 	to [out=-90, in=05] 	(4.30, 1.60) --
		(4.30, 1.60) 	to [out=-175, in=-10] 	(3.30, 1.60) --
		(3.30, 1.60) 	to [out=-190, in=-90] 	(3.00, 1.90);

	\draw[-, dashed, line width=0.35mm] 
		(3.00, 4.58)	to [out=90, in=-175] 	(3.30, 4.90) -- 
		(3.30, 4.90) 	to [out=05, in=-185] 	(5.00, 4.90) -- 
		(5.00, 4.90) 	to [out=-05, in=90] 	(5.50, 3.70) --
		(5.50, 3.70) 	to [out=-90, in=90] 	(5.50, 2.70) --
		(5.50, 2.70) 	to [out=-90, in=05] 	(4.90, 1.60) --
		(4.90, 1.60) 	to [out=-175, in=-05] 	(3.30, 1.60) --
		(3.30, 1.60) 	to [out=-185, in=-90] 	(3.00, 1.90);
	\draw[-, smooth, line width=0.35mm]
		(5.50, 2.70) 	to [out=-90, in=05] 	(4.90, 1.60) --
		(4.90, 1.60) 	to [out=-175, in=-05] 	(3.30, 1.60) --
		(3.30, 1.60) 	to [out=-185, in=-90] 	(3.00, 1.90);

    \node[fill=white,inner sep=1pt] at (4.22, 4.10) {$\hat{\textbf{y}}^{(t-1)}$}; 
    \node[fill=white,inner sep=1pt] at (4.90, 3.30) {$\hat{\textbf{y}}^{(t-2)}$}; 
    \node[fill=white,inner sep=1pt] at (5.55, 2.50) {$\hat{\textbf{y}}^{(t-p)}$}; 

    \node[] at (0.05, 5.6) {\textbf{ARX($p$)}};
    \node[] at (4.05, 5.6) {\textbf{RNN($p$)}};

    \node[] at (0.3, 0.2) {\phantom{---}};

\end{tikzpicture}
  		\caption{\centering{\small{
  		Autoregressive models of order $p$ in the linear \newline
  		(ARX, \textit{left}) and nonlinear (RNN, \textit{right}) cases.}}}
  		\label{fig:sub2}
	\end{subfigure}
	\caption{\small{RNN($p$) models represent the simplest extension of ARX($p$) models, achieved by incorporating an additional nonlinear hidden layer $\vvv{h}{t}$.
	The feedback connections
	--~$\hat{y}^{(t-1)}$ in Figure 1a;  
	$\hat{y}^{(t-1)}$, $\hat{y}^{(t-2)}$, $\hat{y}^{(t-p)}$ in Figure 1b~-- 
	are shown as a dashed line
	followed by a solid line (input to the model at the following time step $t$), 
	indicating that past outputs are used as autoregressive inputs at time $t$.
	The RNN($p$) architecture maintains high degree of interpretability and it is easy to implement and to train.
	}}
	\label{fig:fig1}
\end{figure}

\smallskip
\titoletto{Richiamo ad Algoritmi (Contribution II)}
A central challenge in training RNN($p$) is ensuring that the network effectively learns from the data.
As we discuss in Section \ref{sec:complex},
the key to achieving this is efficient gradient computation, as gradients are essential for updating model parameters during training \citep[see, e.g.,][]{bengio2012practical}.
We study the three main algorithms for gradient computation in general RNNs, developing tailored implementations for RNN($p$) models and computing \textit{analytically} the algorithmic complexities associated with their execution.\footnote{The RNN($p$) architecture is novel and not available in the major deep learning frameworks (TensorFlow and PyTorch), making its efficient implementation and algorithmic analysis a core technical contribution of this paper. 
We provide a complete
implementation of this architecture in a C++ library, available upon request.
}
In particular, we show that the multi-lag structure of RNN($p$) requires careful algorithm selection, as an ill-suited choice can result in significant computational inefficiencies.

\smallskip

{ \color{my_colour2}
We present two applications of RNN($p$) models to power consumption forecasting, a domain of particular relevance to the energy finance sector for several key reasons. Firstly, power consumption forecasting holds significant economic importance, as it directly impacts generation scheduling, portfolio optimisation, and risk management. As widely noted in the literature, accurate forecasts enable electricity providers and market participants to optimise generation, reduce operational costs, and minimise reliance on expensive backup energy sources \citep[see, e.g.,][]{hong2016, hong2020energy}.

Beyond operational efficiency, load forecasts are also fundamental inputs for electricity price formation, forward curve construction, hedging in futures and options markets, and bid optimisation in day-ahead and intraday auctions, where forecast errors result in costly imbalance penalties
\citep[see, e.g.,][]{conejo2010decision, lago2021forecasting, zimmermann2025efficient}.
The link between forecast accuracy and electricity price dynamics has been extensively documented in the literature 
\citep[see, e.g.,][]{hong2016, Nowotarski2018, serafin2022trading}, 
confirming that forecast improvements translate into direct economic value for market participants.
}

Additionally, in energy markets, as in many other fintech applications, the interpretability of predictive models is essential due to the regulatory and economic implications of forecast errors. Transparent architectures such as RNN($p$) provide clear insights that help practitioners understand model behaviour and build trust in the predictions,
supporting both practical implementation and integration into automated systems 
\citep[see, e.g.,][]{fu2021review}.

\titoletto{Application to Load Forecasting}
Through a comprehensive experimental analysis, we show that RNN($p$) models are not only characterised by an elementary design, but they also deliver exceptional predictive performance, even when compared to state-of-the-art deep learning models. We conduct our experiments using two distinct datasets -- one referring to electricity load in the New England region, the other to electricity net load in the London area -- both of which are publicly available to ensure the reproducibility of our findings.

\medskip
\titoletto{Contributions}
The main contributions of this paper are threefold.
First, we introduce a new elementary architecture for Recurrent Neural Networks, the RNN($p$), which is particularly suited to problems with inherent seasonalities and to integrating expert-informed features.
Second, we analyse the main learning algorithms for RNN($p$), providing an analytical computation of their leading-order time and space complexities.
Third, we present two applications of these models to power consumption forecasting, showing that they can achieve excellent accuracy in both load and net load forecasting.

\titoletto{The rest of the paper}
The rest of the paper is organised as follows. 
In Section \ref{sec:model_spec}, we introduce the modelling framework and define the RNN($p$) architecture. 
In Section \ref{sec:complex}, we describe the main learning algorithms for RNN($p$), providing tailored implementations and deducing their time and space complexities.
In Section \ref{sec:energy}, we conduct an empirical evaluation of RNN($p$) models in the context of power consumption forecasting.
Finally, Section \ref{sec:conclusions} summarises the results and states our conclusions.

\section{RNN($p$): Model Description} \label{sec:model_spec}

\titoletto{Cappellino}
In this section, we provide a detailed description of the RNN($p$) architecture, outlining its mathematical structure and how it is designed to capture multiple seasonalities. Additionally, we
provide a concise overview of related research
to frame the model within the existing literature.

\smallskip
\titoletto{The RNN(p) model: equation}
An RNN($p$) model is a Neural Network with one hidden layer and $p$ lagged feedbacks of Jordan type, i.e.\
feedback connections from the output layer to the hidden layer 
\citep[cf.][]{jordan1986}.
Given an input sequence $\{ \vvv{x}{t} \}_{t=1}^{\tau}$ of length $\tau$,
the model is defined by the following equations, for $t = 1, \dots, \tau$:
\begin{subnumcases}{\label{eq:RNN_AB}}
	\displaystyle
	\, ~
	\vphantom{\sum_{k=1}^{p}}
	\vvv{h}{t}
	=
	\actf
	\left(
		\vv{b} +
		U \vvv{x}{t} +
		\sum_{k=1}^{p} W_i ~ \vvh{y}{t-k}
	\right) 
	\label{eq:RNN_A}
	\\[-2mm]
	\, ~
	\vphantom{\sum_{k=1}^{p}}
	\vvh{y}{t}
	=
	\vv{c} +
	V \vvv{h}{t} 
	\,
	\label{eq:RNN_B}
\end{subnumcases}
where $\vvv{x}{t} \in \erre{x}$
is the vector of exogenous variables at time $t$,
$\vvv{h}{t} \in \erre{h}$
the hidden state, and
$\vvh{y}{t} \in \erre{y}$
the output of the model.\footnote{We adopt a notation that closely follows the reference textbook by \citet[][cf., e.g., p.381]{goodfellow2016}.}
For $t \leq 0$, the output is initialised as 
$\vvh{y}{t} = (0, \, \dots, \, 0)^\top \in \erre{y}$.

The recurrence in \eqref{eq:RNN_A} is defined on multiple lagged outputs
$\{\vvh{y}{t-k}\}_{k=1}^p$, rather than only on the previous hidden state
$\vvv{h}{t-1}$ as in Elman-type RNNs \citep[][]{elman1990}, which are pre-implemented in many deep learning frameworks. 
By feeding back past outputs, RNN($p$) implements an output-driven recurrence, yielding an autoregressive structure that closely mirrors the well-known ARX($p$) models and enables explicit modelling of longer-term dependencies.

\smallskip
In our model, the hidden state $\vvv{h}{t}$ is obtained by applying the nonlinear activation function $\actf:\erre{h} \rightarrow \erre{h}$ to
the \textit{pre-activation} state
\begin{equation*}
	\vvv{a}{t}
	:=
	\vv{b} +
		U \vvv{x}{t} +
		\sum_{k=1}^{p} W_i ~ \vvh{y}{t-k} \,,
\end{equation*}
as shown in \eqref{eq:RNN_A};
as common in deep learning, 
we consider activation functions
$\actf(\boldsymbol{\cdot})$ that are applied
component-wise to the vector
$\vvv{a}{t}$, such as the well-known sigmoid and ReLU functions
\citep[see, e.g.,][]{dubey2022activation}.
Both vectors $\vvv{a}{t}$ and $\vvv{h}{t}$ have dimension $h$,
which is, as standard in the literature,
set larger than both the output dimension $y$ and the order $p$ of the model to ensure high representation capacity
\citep[see, e.g.,][]{bengio2012practical}.

\smallskip
The matrices 
$U \in \erre{h \times x}$,\, $W_1, \dots, W_p \in \erre{h \times y}$
and the vector $\vv{b} \in \erre{h}$ 
are the trainable parameters (called the \textit{weights}, hereinafter) of the input-to-hidden map \eqref{eq:RNN_A};
the matrix
$V \in \erre{y \times h}$
and
the vector
$\vv{c} \in \erre{y}$
are the weights of the hidden-to-output map \eqref{eq:RNN_B}.
To simplify the notation, we flatten and collect all input-to-hidden weights in a single vector $\boldtheta$ such that
\begin{equation}
	\boldtheta := 
	\left\{ 
		U, \left\{W_i\right\}_{i=1}^{p}, \vv{b}
	\right\}
	\,
	\in \erre{\vartheta} \,,
\end{equation}
whose dimension is 
$\vartheta := (x + py + 1)h$,
and all hidden-to-output weights in a single vector $\boldphi$ such that
\begin{equation}
	\boldphi := 
	\left\{ 
		V, \vv{c}
	\right\}
	\,
	\in \erre{\varphi} \,,
\end{equation}
whose dimension is $\varphi:= (h+1)y$.
Thus, the total number of weights in an RNN($p$) is
\begin{equation} \label{eq:wtotalweights}
	w 
	:= \vartheta + \varphi
	 = (x + py + 1)h + (h+1)y
	 \,.
\end{equation}

\titoletto{Nonlinear equation and activation function}
Input data for RNN($p$) models take the form of sequences of vectors. As standard in forecasting applications \citep[see, e.g.,][]{goodfellow2016}, we consider in our analysis input sequences with a fixed length $\tau$,
corresponding to the lookback period used to predict future values.
Hence, each input sequence is an ordered collection of vectors 
$\{ \vvv{x}{t} \}_{t=1}^{\tau}$. 

\smallskip
\titoletto{Tabella dei parametri}
Table \ref{tab:notation} summarises the dimensions 
of the considered vectors, together with the main relevant quantities introduced in this section.
These will be crucial for the computation of time complexity and space complexity of the learning algorithms in Section \ref{sec:complex}.

\begin{table}[!hb]
\centering
\begin{tabular}{c l || c l}
   \bottomrule 
    $\phantom{-}p\phantom{-}$ & 
    	Number of feedback connections ~ &
	$\phantom{-}\vartheta\phantom{-}$ & 
		Number of input-to-hidden weights \\
	$x$ & 
		Dimension of exogenous input vector &
	$\varphi$ & 
		Number of hidden-to-output weights ~ \\
	$h$ & 
		Number of hidden neurons &
	$w$ & 	
		Total number of weights \\
	$y$ & 
		Dimension of output vector &
	$\tau$ & 
		Length of input sequence \\
    \toprule
\end{tabular}
\caption{Key dimensions and symbols in RNN($p$) models.}
\label{tab:notation}
\end{table}

\titoletto{Literature}
In the context of existing research, RNN($p$) models can be classified as part of 
the broad class of nonlinear autoregressive models with exogenous inputs
\citep[NARX models; see, e.g.,][]{billings2013}.
This class encompasses a wide variety of models, arising from the many different ways nonlinearity
can be introduced -- for example, through
polynomial bases \citep[see, e.g.,][]{piroddi2003identification},
wavelet functions \citep[see, e.g.,][]{wei2006long}
or Neural Networks.
Of particular relevance to the present work is the latter category of models;
notable references in the literature are the early studies on multilayer networks 
\citep[see, e.g.,][]{narendra1990identification, siegelmann1997computational},
as well as more recent works on Layered Digital Dynamic Networks \citep[LDDNs; cf.][]{dejesus2007, nguyen2011stability, beale2024}.

\titoletto{Controcappello e novità v1}
While these existing models are designed for general-purpose versatility -- through multilayer compositions and flexible feedback connections -- the RNN($p$) family is specifically tailored to address the challenge of inherent seasonalities on multiple time scales.
Rather than relying on deep architectures, RNN($p$) models adopt an elementary autoregressive structure,
as illustrated in Figure \ref{fig:fig1}, representing the simplest nonlinear extensions of ARX($p$) models.
The advantages of this design are several:
the explicit use of lagged feedback supports interpretability, in line with the ARX($p$) framework; 
the model retains the ability to learn complex nonlinear dependencies between inputs and outputs \citep[as guaranteed by the Universal Approximation Theorem; see, e.g.,][]{goodfellow2016}; 
and finally, 
the training procedure is made straightforward and computationally efficient.

In the next section, we focus specifically on this last aspect, analysing the learning algorithms that can be used to train these models effectively.

\section{Learning Algorithms for RNN($p$)}  \label{sec:complex}

\titoletto{Cappellino: Motivation}
The performance of an RNN($p$) largely depends on its ability to learn meaningful patterns from the data.
The training of an RNN($p$) is mathematically framed as an optimisation problem:
given the observed time series $\{\vvv{y}{t}\}_{t}$ and the corresponding network outputs
$\{\vvh{y}{t}\}_{t}$, the model weights $\boldtheta$ and $\boldphi$ are estimated by minimising
\begin{equation}
\mathcal{L}(\boldtheta,\boldphi)
:=
\sum_{t}
L\!\left(\vvv{y}{t}, \vvh{y}{t}(\boldtheta,\boldphi)\right) \,,
\label{eq:minimisation_problem}
\end{equation}
where $L$ is the (per-sample) loss function 
--~that quantifies the error between each predicted value $\vvh{y}{t}$ and the true value $\vvv{y}{t}$~-- 
and $\mathcal{L}(\boldtheta,\boldphi)$ is the aggregate loss over all samples.

Neural Networks are commonly trained using gradient-based optimisers, which are designed to iteratively adjust the model weights to reduce the aggregate loss $\mathcal{L}(\boldtheta,\boldphi)$ in \eqref{eq:minimisation_problem}, and thus improve predictive performance \citep[see, e.g.,][]{goodfellow2016, aggrawal2018}.
These optimisers require the computation of the gradients of $\mathcal{L}$ with respect to 
$\boldtheta$ and $\boldphi$.
In the context of the RNN($p$) models introduced in this paper, 
this involves first computing the gradients of the (per-sample) loss function $L$\,:
\begin{equation}
	\dv{\loss}{\boldtheta} \,
	\in
	\erre{1 \times \vartheta}
	\hspace{0.7cm}
	\text{and}
	\hspace{0.7cm}
	\dv{\loss}{\boldphi} \,
	\in 
	\erre{1 \times \varphi}
	\hspace{0.15cm}
	\,.
	\label{eqn:gradient_000}
\end{equation}
The gradients of $\mathcal{L}(\boldtheta,\boldphi)$ are then obtained by summing these quantities over the whole time series.

Unlike FNNs, for which the gradients can be easily obtained, the recurrent nature of RNN($p$) models makes this computation non-trivial.
In an RNN($p$), the predicted output $\vvh{y}{\tau}$ depends not only directly on the weights $\boldtheta$ and $\boldphi$ -- as seen, respectively, in \eqref{eq:RNN_A} and \eqref{eq:RNN_B} -- but also indirectly through the previous outputs, which in turn are functions of those same weights.
As a result, the derivative of the loss must capture both direct dependencies (i.e.\ \textit{partial} derivatives) and recurrent dependencies, requiring the computation of \textit{total} derivatives. These account for all pathways through which the model weights $\boldtheta$ and $\boldphi$ influence the output, whether immediately or through the recurrent structure of the model.

\smallskip
\titoletto{What's new?}
In this section, we illustrate how the three principal algorithms for gradient computation for Neural Networks can be adapted to
RNN($p$) models.\footnote{
Additional details on the implementation of the algorithms are provided in Appendix \ref{sec:appendixALGO}.
}
Leveraging the elementary RNN($p$) architecture, we derive for each learning algorithm:
i) closed-form analytical expressions for the gradient;
ii) accurate leading-order estimates of both time complexity and space complexity.
These results constitute a core theoretical contribution of the paper, enabling a rigorous comparison of the learning algorithms within the RNN($p$) framework.

\subsection{The algorithms} \label{sec:thealgo}

Real-Time Recurrent Learning (RTRL), Backpropagation Through Time (BPTT), and Adjoint Automatic Differentiation (AAD) are the three main algorithms considered in the literature for computing gradients in general RNNs \citep[see, e.g.,][]{goodfellow2016, baydin2018}.
Each of them relies on a distinct decomposition method to obtain the gradient of the loss function with respect to the network weights.
Nevertheless, it is important to note that
all three algorithms perform exact gradient computation,
resulting in identical outputs across the methods.

\smallskip
In the following, we detail the adaptation of these three algorithms to the RNN($p$) models.
When training an RNN($p$), the full time series is partitioned into shorter sequences of length $\tau$. For each learning algorithm, we deduce the gradient of the loss $L$ for a single such sequence.
An important term that appears in all cases is the derivative, at time $t$, of the nonlinear transformation $\vvv{h}{t} = \actf(\vvv{a}{t})$; we denote this as the matrix
\begin{equation} \label{eq:A_t}
	A^{(t)} := \dv{\vvv{h}{t}}{\vvv{a}{t}} \in \erre{h \times h} \,.
\end{equation}
Since we consider, as standard in the literature, activation functions which act component-wise, the matrix $A^{(t)}$ is diagonal for every $t = 1, \dots, \tau$.

\subsubsection{RTRL}
RTRL is a learning algorithm grounded in the early work of
\citet{williams1989} and later refined in subsequent developments \citep[see, e.g.,][and references therein]{caterini2018recurrent}.
In the context of RNN($p$) models, we can formulate this algorithm as follows.

First, we compute the total derivatives of the output $\vvh{y}{t}$ with respect to the model weights, i.e.\ $\dvinline{\vvh{y}{t}}{\boldtheta}$ and $\dvinline{\vvh{y}{t}}{\boldphi}$, for each $t = 1, \dots, \tau$. These derivatives are progressively calculated as the network runs forward, from the initial time step ($t=1$) to the final time step ($t=\tau$), using the relationships:
\begin{subnumcases}{\label{eq:RNN_RTRL_ab}}
	\displaystyle
	\, ~
	\dv{\vvh{y}{t}}{\boldtheta}
	=
	V
	A^{(t)}
	\left(
		\pdv{\vvv{a}{t}}{\boldtheta}
		+
		\sum_{k=1}^{\min(p, t-1)}
		W_k
		\dv{\vvh{y}{t-k}}{\boldtheta}
	\right) 
	\,,
	\label{eq:RNN_RTRL2a} 
	\\
	\, ~
	\dv{\vvh{y}{t}}{\boldphi}
	=
	\pdv{\vvh{y}{t}}{\boldphi}
	+
	V
	A^{(t)} \,
	\sum_{k=1}^{\min(p, t-1)}
	 	W_k
		\dv{\vvh{y}{t-k}}{\boldphi}
		\hspace{1mm}.
	\label{eq:RNN_RTRL2b}
\end{subnumcases}

Then, once the final total derivatives $\dvinline{\vvh{y}{\tau}}{\boldtheta}$ and $\dvinline{\vvh{y}{\tau}}{\boldphi}$ have been obtained, we compute the two gradients in \eqref{eqn:gradient_000} as:
\begin{equation}
	\displaystyle
	\, ~
	\mygradient{\loss}{\boldtheta}
	=
	\pdv{\loss}{ \vvh{y}{\tau} }
	~
	\dv{ \vvh{y}{\tau} }{\boldtheta} \,
	\hspace{0.7cm}
	\text{and}
	\hspace{0.7cm}
	\, ~
	\mygradient{\loss}{\boldphi}
	=
	\pdv{\loss}{ \vvh{y}{\tau} }
	~
	\dv{ \vvh{y}{\tau} }{\boldphi} \,
\end{equation}
where $\pdvinline{\loss}{ \vvh{y}{\tau} }$ denotes the derivative of loss function with respect to the final output $\vvh{y}{\tau}$.

\smallskip
As we observe in \eqref{eq:RNN_RTRL_ab}, at every time $t$, the total derivatives
$\dvinline{\vvh{y}{t}}{\boldtheta}$ and 
$\dvinline{\vvh{y}{t}}{\boldphi}$ 
are calculated as a function of: 
i) their own previous $p$ values, and
ii) the \textit{partial} derivatives of the output $\vvh{y}{t}$ with respect to $\boldtheta$ and $\boldphi$, which can be easily deduced from \eqref{eq:RNN_AB}.
This structural simplicity is the key advantage of RTRL:
it makes the algorithm easy to implement, and allows it to operate without storing previous inputs or hidden states of the network -- only the most recent $p$ total derivatives are needed.

Conversely, the main challenge of RTRL lies in the matrix operations involved, particularly the repeated multiplications between the weight matrices
$W_k \in \mathbb{R}^{h \times y}$
and the derivative matrices
$\dvinline{\vvh{y}{t-k}}{\boldtheta} \in \mathbb{R}^{y \times \vartheta}$
and
$\dvinline{\vvh{y}{t-k}}{\boldphi} \in \mathbb{R}^{y \times \varphi}$.
These operations can become computationally demanding as the network size increases, that is, when the number of neurons or the number of feedbacks grows.

\subsubsection{BPTT} \label{ssec:bptt}

It is well known that vector-matrix multiplications are significantly more efficient than matrix-matrix multiplications.
Transforming the computation of the gradient into a series of vector-matrix operations is the premise of BPTT \citep[][]{werbos1990}, the industry standard algorithm for training RNNs.
This algorithm operates by \textit{unrolling} the recurrent structure of the RNN over time, effectively converting it into a FNN \citep[see, e.g.,][p.378]{goodfellow2016}. 
The gradient computation is thus achieved by: i) processing the entire sequence forward, storing all inputs, hidden states and outputs; ii) propagating the gradient through the unrolled network, backward in time.

\smallskip
In the standard case of a single-lag RNN, i.e.\ RNN($1$), the unrolling procedure results in a chain-like computational graph, as illustrated in Figure \ref{fig:figure_bptt}a \citep[see also, e.g.,][]{pascanu2013}.
Within this structure, the gradient of the loss function with respect to model weights is computed by applying the chain rule of derivatives backward in time.
In particular, the gradients in an RNN($1$) are unfolded as:
\begin{subnumcases}{\label{eq:RNN_BPTT_ab}}
	\displaystyle
	\, ~
	\dv{\loss}{ \boldtheta }
	\,=\,
	\pdv{\loss}{ \vvh{y}{\tau} }	
	\,
	\dv{\vvh{y}{\tau}}{\boldtheta}
	\overset{\small{\eqref{eq:RNN_RTRL2a}}}{\,\,=\,\,}
	\pdv{\loss}{ \vvh{y}{\tau} }
	V
	A^{(\tau)}
	\left(
		\pdv{\vvv{a}{\tau}}{\boldtheta}
		+
		W_1
		\dv{\vvh{y}{\tau-1}}{\boldtheta}
	\right)
	\overset{\small{\eqref{eq:RNN_RTRL2a}}}{\,\,=\,\,}
	\cdots
	\label{eq:RNN_BPTT_1a}
	\\
	\displaystyle
	\, ~
	\dv{\loss}{ \boldphi }
	\, = \,
		\pdv{\loss}{ \vvh{y}{\tau} }
		\pdv{\vvh{y}{\tau}}{\boldphi}
	\overset{\small{\eqref{eq:RNN_RTRL2b}}}{\,\,=\,\,}
	\pdv{\loss}{ \vvh{y}{\tau} }
	\left(
		\pdv{\vvh{y}{\tau}}{\boldphi}
		+
		V
		A^{(\tau)}
		 W_1
		\dv{\vvh{y}{\tau-1}}{\boldphi}
	\right)
	\overset{\small{\eqref{eq:RNN_RTRL2b}}}{\,\,=\,\,}
	\cdots
	\label{eq:RNN_BPTT_1b}
\end{subnumcases}
These expressions are obtained by recursively applying the gradient identities in
\eqref{eq:RNN_RTRL_ab}, moving backward from the final time step $(t=\tau)$ to the initial step
$(t=1)$.

\smallskip
\titoletto{BPTT: The problem with RNN(p)}
Nevertheless, BPTT becomes significantly more complex when extended to general RNN($p$) models with $p>1$. 
In this setting, the presence of multiple feedbacks introduces nested temporal dependencies, which give rise to nested summations in the gradient expressions \eqref{eq:RNN_BPTT_ab}. This drastically increases the computational workload of the backpropagation procedure.

The increase in algorithmic complexity is reflected in the structure of the unrolled computational graph. Figure \ref{fig:figure_bptt}b shows the computational graph for an RNN($2$): as every output 
$\vvh{y}{t}$ depends on the two previous outputs $\vvh{y}{t-1}$ and $\vvh{y}{t-2}$, the unrolling generates a binary tree, in contrast to the linear chain seen in the RNN($1$) case.

\begin{figure*}[!b]
    \begin{minipage}[b]{0.99\textwidth}
        \centering
		\resizebox{\linewidth}{!}{
\begin{tikzpicture}[scale=.9, ->, thick, draw=black!75, shorten <= 5pt, shorten >= 5pt]
        
    \draw (16,4.5) circle(0.0) node[] (L) {\small{$\loss$}};
    \draw[rounded corners] (15.15, 4.15) rectangle ++(1.7, 0.7) {};
           
    \draw (7,2.5) circle(0.0) node[] (Y-2) {$\vvh{y}{1}$};
    \draw[rounded corners] (6.15, 2.15) rectangle ++(1.7, 0.7) {};
    \draw (7,1) circle(0.0) node[] (A-2) {$\vvv{h}{1}$};
    \draw[rounded corners] (6.15, 0.65) rectangle ++(1.7, 0.7) {};         
  
    \draw (10,2.5) circle(0.0) node[] (Y-3) {$\vvh{y}{\mathbf{\cdot}}$};
    \draw[rounded corners] (9.15, 2.15) rectangle ++(1.7, 0.7) {};
    \draw (10,1) circle(0.0) node[] (A-3) {$\vvv{h}{\mathbf{\cdot}}$};
    \draw[rounded corners] (9.15, 0.65) rectangle ++(1.7, 0.7) {};   
            
    \draw (13,2.5) circle(0.0) node[] (Y-4) {$\vvh{y}{\tau-1}$};
    \draw[rounded corners] (12.15, 2.15) rectangle ++(1.7, 0.7) {};
    \draw (13,1) circle(0.0) node[] (A-4) {$\vvv{h}{\tau-1}$};
    \draw[rounded corners] (12.15, 0.65) rectangle ++(1.7, 0.7) {};   
    
    \draw (16,2.5) circle(0.0) node[] (Y-5) {$\vvh{y}{\tau}$};
    \draw[rounded corners] (15.15, 2.15) rectangle ++(1.7, 0.7) {};
    \draw (16,1) circle(0.0) node[] (A-5) {$\vvv{h}{\tau}$};
    \draw[rounded corners] (15.15, 0.65) rectangle ++(1.7, 0.7) {};

	\draw[->, CadetBlue, dashed] (7, 2.2) -- (7, 1.3);
    \draw[CadetBlue, dashed] (7,1.9) circle(0.0) node[right] {};

	\draw[->, CadetBlue, dashed] (10, 2.2) -- (10, 1.3);
    \draw[CadetBlue, dashed] (10,1.9) circle(0.0) node[right] {};

	\draw[->, CadetBlue, dashed] (12.9, 2.2) -- (12.9, 1.3);
    \draw[CadetBlue, dashed] (12.9,1.9) circle(0.0) node[right] {};

	\draw[->, CadetBlue, dashed] (16, 2.2) -- (16, 1.3);
    \draw[CadetBlue, dashed] (16,1.9) circle(0.0) node[right] {};

	\draw[->, CadetBlue, dashed] (16, 4.3) -- (16, 2.8);
    \draw[CadetBlue, dashed] (16,4) circle(0.0) node[right] {};

	\draw[->, CadetBlue, dashed] (15.20, 1.5) -- (13.9, 2.3);
	\draw[->, CadetBlue, dashed] (12.20, 1.5) -- (10.9, 2.3);
	\draw[->, CadetBlue, dashed] ( 9.20, 1.5) -- ( 7.9, 2.3);

    \draw[CadetBlue, dashed] ( 9.9,2.6) circle(0.0) node[right] {};
    \draw[CadetBlue, dashed] (13.9,2.6) circle(0.0) node[right] {};

\begin{scope}[xshift=16.25cm, yshift=-13.5cm]
    
    \draw (8,18) circle(0.0) node[] (L) {\small{$\loss$}};
    \draw[rounded corners] (7.15, 17.65) rectangle ++(1.7, 0.7) {};
    
    \draw (8,16.0) circle(0.0) node[] (AY-1) {$\vvh{y}{\tau}$};
    \draw[rounded corners] (7.15, 15.65) rectangle ++(1.7, 0.7) {};
    \draw (8,14.5) circle(0.0) node[] (A-1) {$\vvv{h}{\tau}$};
    \draw[rounded corners] (7.15, 14.15) rectangle ++(1.7, 0.7) {};

    \draw [] (5,13)  circle (0.0) node[] (BY-1) {$\vvh{y}{\tau-1}$};
    \draw[rounded corners] (4.15, 12.65) rectangle ++(1.7, 0.7) {};
    \draw [] (5,11.5)  circle (0.0) node[] (B-1) {$\vvv{h}{\tau-1}$};
    \draw[rounded corners] (4.15, 11.15) rectangle ++(1.7, 0.7) {};
    
    \draw [] (11,13) circle (0.0) node[] (BY-2) {$\vvh{y}{\tau-2}$};
    \draw[rounded corners] (10.15, 12.65) rectangle ++(1.7, 0.7) {};
    \draw [] (11,11.5) circle (0.0) node[] (B-2) {$\vvv{h}{\tau-2}$};
    \draw[rounded corners] (10.15, 11.15) rectangle ++(1.7, 0.7) {};

    \draw [] (3.50,9.5)  circle (0.0) node[] (CY-1) {$\vvh{y}{\tau-2}$};
    \draw[rounded corners] (2.65, 9.15) rectangle ++(1.7, 0.7) {};
    \draw [] (6.50,9.5)  circle (0.0) node[] (CY-2) {$\vvh{y}{\tau-3}$};
    \draw[rounded corners] (5.65, 9.15) rectangle ++(1.7, 0.7) {};
    \draw [] (9.50,9.5) circle (0.0) node[] (CY-3) {$\vvh{y}{\tau-3}$};
    \draw[rounded corners] (8.65, 9.15) rectangle ++(1.7, 0.7) {};
    \draw [] (12.50,9.5) circle (0.0) node[] (CY-4) {$\vvh{y}{\tau-4}$};
    \draw[rounded corners] (11.65, 9.15) rectangle ++(1.7, 0.7) {};

    \draw [] (3.50,8)  circle (0.0) node[] (C-1) {$\vvv{h}{\tau-2}$};
    \draw[rounded corners] (2.65, 7.65) rectangle ++(1.7, 0.7) {};
    \draw [] (6.50,8)  circle (0.0) node[] (C-2) {$\vvv{h}{\tau-3}$};
    \draw[rounded corners] (5.65, 7.65) rectangle ++(1.7, 0.7) {};
    \draw [] (9.50,8) circle (0.0) node[] (C-3) {$\vvv{h}{\tau-3}$};
    \draw[rounded corners] (8.65, 7.65) rectangle ++(1.7, 0.7) {};
    \draw [] (12.50,8) circle (0.0) node[] (C-4) {$\vvv{h}{\tau-4}$};
    \draw[rounded corners] (11.65, 7.65) rectangle ++(1.7, 0.7) {};

    \path[CadetBlue, dashed] (L) edge[] node 
        [right] {} (AY-1);
    \path[CadetBlue, dashed] (AY-1) edge[] node 
        [right] {} (A-1);

    \path[CadetBlue, dashed] (A-1) edge[] node [above left] {} (BY-1);
    \path[CadetBlue, dashed] (A-1) edge[] node [above right] {} (BY-2);

    \path[CadetBlue, dashed] (BY-1) edge[] node 
        [right] {} (B-1);
    \path[CadetBlue, dashed] (BY-2) edge[] node 
        [left] {} (B-2);

    \path[CadetBlue, dashed] (B-1) edge[] node 
    	[above left= 0.1cm,anchor=320] {} (CY-1);
    \path[CadetBlue, dashed] (B-1) edge[] node 
    	[above right= 0.1cm, anchor=220] {} (CY-2);
    \path[CadetBlue, dashed] (B-2) edge[] node 
    	[above left= 0.1cm, anchor=320] {} (CY-3);
    \path[CadetBlue, dashed] (B-2) edge[] node 
    	[above right= 0.1cm,anchor=220] {} (CY-4);

    \path[CadetBlue, dashed] (CY-1) edge[] node [right] {} (C-1);
    \path[CadetBlue, dashed] (CY-2) edge[] node [left] {} (C-2);
    \path[CadetBlue, dashed] (CY-3) edge[] node [right] {} (C-3);
    \path[CadetBlue, dashed] (CY-4) edge[] node [left] {} (C-4);

\end{scope}
 
    \node[] at ( 7.2,  4.50) {\textit{(a)} \textbf{RNN($1$)}};
    \node[] at (20.0,  4.50) {\textit{(b)} \textbf{RNN($2$)}};
 
\end{tikzpicture}
       }
    \end{minipage}
 	\vspace{0.4cm}
	\caption{\small{
	Gradient backpropagation in an RNN($1$) (\textit{left}) and in an RNN($2$)
	(\textit{right}).
In the RNN($1$) case, the unrolling procedure results in a linear chain, with each output $\vvh{y}{t}$ depending only on the previous output $\vvh{y}{t-1}$. In the RNN($2$) case, each output $\vvh{y}{t}$ depends on the two previous outputs, $\vvh{y}{t-1}$ and $\vvh{y}{t-2}$, resulting in a binary tree structure.
	}}
	\label{fig:figure_bptt}
\end{figure*}

Backpropagating through such a tree is computationally expensive, since the overall cost scales with the number of nodes in the tree. As we prove in Section \ref{ssec:tcsc}, when $p > 1$, this number grows exponentially with the sequence length 
$\tau$, resulting in an exponential increase in the time required to compute gradients. 

\subsubsection{AAD}

It is possible to solve the computational challenges posed by BPTT using AAD
\citep[see, e.g.,][and references therein]{baydin2018}.
By exploiting the adjoint formulation of the total derivatives, this algorithm avoids the need to explicitly unroll the network over time, allowing gradients to be computed more efficiently.
Like BPTT, AAD operates backward in time and requires the entire sequence to be processed before the gradient can be assembled.
In the RNN($p$) framework, we can formulate AAD as follows.

First, we compute for each $t$ the total derivative of the loss with respect to the output $\vvh{y}{t}$, i.e.\ 
$\dvinline{L}{\vvh{y}{t}}$. These derivatives are progressively calculated from the final time step ($t=\tau$) to the initial time step $(t=1)$, using the relationships:
\begin{subnumcases}{\label{eq:aad_ab}}
	\displaystyle
	\, ~
	\dv{L}{\vvh{y}{t}} =
		\pdv{L}{\vvh{y}{\tau}}
		& \textnormal{if} \, $t = \tau$
	\\[0.1cm]
	\, ~
	\dv{L}{\vvh{y}{t}} =
	\sum_{k=1}^{\min(p, \tau-t)}
		\dv{L}{\vvh{y}{t+k}} \,
		V A^{(t+k)} W_k
		\hspace{0.5cm}
		& \textnormal{if} \, $t = \tau-1, \dots, 1$
\end{subnumcases}
In this case, each total derivative $\dvinline{L}{\vvh{y}{t}}$ is computed as a function of its own subsequent $p$ values.

Once all total derivatives $\dvinline{L}{\vvh{y}{t}}$ have been obtained, we multiply them by the partial derivatives of the outputs with respect to the weights, i.e.\ $\pdvinline{ \vvh{y}{t} }{\boldtheta}$ and $\pdvinline{ \vvh{y}{t} }{\boldphi}$. The gradients in \eqref{eqn:gradient_000} are finally computed by summing the contributions across all time steps:
\begin{equation}
	\displaystyle
	\, ~
	\dv{\loss}{\boldtheta}
	=
	\sum_{t=1}^{\tau}
		\dv{\loss}{ \vvh{y}{t} }
		~
		\pdv{ \vvh{y}{t} }{\boldtheta} \,
	\hspace{0.7cm}
	\text{and}
	\hspace{0.7cm}
	\, ~
	\dv{\loss}{\boldphi}
	=
	\sum_{t=1}^{\tau}
		\dv{\loss}{ \vvh{y}{t} }
		~
		\pdv{ \vvh{y}{t} }{\boldphi}
\end{equation}

This compact reconstruction of the gradients, achieved by iteratively aggregating contributions backward in time, is the key insight of AAD. This algorithm minimises redundancy in the computation -- by reusing shared terms across time steps rather than recomputing them -- effectively generalising backpropagation to RNNs with multiple feedbacks.
The resulting computational improvements are extremely significant, as we discuss below.

\subsection{Time Complexity and Space Complexity} \label{ssec:tcsc}

In this section, we examine the computational efficiency of the three algorithms introduced in Section \ref{sec:thealgo}, focusing on their algorithmic complexities.
A clear understanding of these aspects is essential for the effective training of RNN($p$) models, especially when working with large datasets and long sequences.

The following proposition represents the core analytical result of the paper, providing explicit expressions for both time complexity and space complexity of the learning algorithms.
Notably, one of these expressions includes a term related to the $p$-bonacci sequences, a generalisation of the well-known Fibonacci sequence; further details on $p$-bonacci numbers, including their definition and key properties, are available in Appendix \ref{sec:appendixD}.

\begin{proposition} \label{prop:complexities}
Given an input sequence of length $\tau$, the gradient computation in an \textnormal{RNN($p$)} model presents the following time and space complexities:

\begin{center}
\begin{tabular}{C{1.5cm} | L{4cm} L{4cm}}
    \toprule
    & ~\textnormal{Time Complexity} & \textnormal{Space Complexity} \\
    \midrule
    \textnormal{AAD} \vphantom{\big[} & 
    	~~~$\mathcal{O}(\tau\, h\, w)$ & 	
    	~$\mathcal{O}(\tau\, (x+h+y) )$ \\[0.05cm]
    \textnormal{RTRL} \vphantom{\big[}& 
    	~~~$\mathcal{O}(\tau\, p\, y\, h\, w)$ & 	
    	~$\mathcal{O}(p\, y\, w)$ \\[0.05cm]
    \textnormal{BPTT} \vphantom{\big[}& 
    	~~~$\mathcal{O}(S_{\tau, \, p}\, h\, w)$ & 
    	~$\mathcal{O}(\tau\, (x+h+y) )$ \\[0.05cm]
    \midrule
\end{tabular}
\end{center}
Here, $h, p, w, x, y$ denote the key model dimensions in Table \ref{tab:notation}, and $S_{\tau, \, p}$ the sum of the first $\tau$ terms of the $p$-bonacci sequence. This sum grows exponentially with $\tau$ when $p\geq2$,  as detailed in Lemma \ref{lemma:p-bonacci}.
\end{proposition}
\begin{proof}
	See Appendix \ref{sec:appendixA}.
\end{proof}

\noindent
The proposition highlights substantial differences in the efficiency of the three learning algorithms.

\smallskip
\titoletto{Time Complexities}
In terms of \textit{time complexity}, AAD proves to be the most efficient algorithm, with a complexity proportional to the product of sequence length $\tau$, the number of hidden neurons $h$, and the total number of trainable weights $w$. Notably, the number of feedback connections $p$ does not appear in the leading-order term, contributing only a lower-order effect to the overall computational cost.

In the special case $p=1$ -- i.e.\ for the RNN($1$) commonly found in reference textbooks \citep[see, e.g.,][]{goodfellow2016} -- the two algorithms BPTT and AAD coincide,
yielding the same time complexity. 
This corresponds to the case in which the unrolling procedure produces a simple chain-like structure, as discussed in Section \ref{sec:thealgo}. 
However, for $p \geq 2$, the time complexity of BPTT grows exponentially with $\tau$ due to the $p$-bonacci term $S_{\tau, p}$ (cf.\ Lemma \ref{lemma:p-bonacci}), making the algorithm computationally prohibitive for anything beyond very short sequences.

An intermediate computational behaviour is observed in the case of RTRL. Like AAD, its time complexity scales linearly with $\tau$; however, it also includes a multiplicative dependence on both the number of feedbacks
$p$ and the output dimension
$y$, resulting in a leading-order cost that is approximately
$py$ times higher than that of AAD.
As a result, when a large number of feedback connections $p$ is involved, or when the output dimension $y$ is greater than one, AAD can become significantly faster than RTRL.

\smallskip
\titoletto{Space Complexities}
In terms of \textit{space complexity}, RTRL is the algorithm that achieves the best performance. 
As previously discussed, RTRL only requires the storage of the total derivatives for the previous $p$ time-steps, making it more memory-efficient than the other two learning algorithms.
In contrast, both AAD and BPTT require the storage of all inputs, hidden states and outputs of the network for every $t = 1, \dots, \tau$, yielding the same leading-order space complexity.
However, in practical applications, the space complexity of all three algorithms is generally comparable, and it never poses execution issues when using RNN($p$) for time series forecasting.

\bigskip
\titoletto{Conclusione della sezione}
In this section, we have derived the key algorithmic tools needed to train RNN($p$) models, including explicit gradient expressions and accurate estimates of computational cost. Moreover, in Proposition \ref{prop:complexities}, we have proved the significant added value of AAD over the other two learning algorithms, particularly in handling multiple feedbacks efficiently. The upcoming section shifts from theoretical analysis to a practical evaluation of RNN($p$), examining their predictive performance in a challenging forecasting task.

\section{Application to Power Consumption Forecasting}  \label{sec:energy}

\titoletto{Cappellino}
This section presents an in-depth experimental analysis of RNN($p$) models applied to power consumption forecasting. 
Power consumption provides an example of a time series with an inherent temporal structure, due to the interplay of natural phenomena and human behaviour, making it a natural setting for the application of these models.
Furthermore, power consumption forecasting holds significant financial importance, as accurate forecasts directly affect operational costs, particularly imbalance costs, and support critical decision-making processes  \citep[see, e.g.,][]{hong2020energy}.

\titoletto{Mid-term and Hourly}
We focus on mid-term power consumption forecasting, with an hourly resolution and a one-year-ahead forecasting horizon.
Hourly forecasts offer the standard temporal granularity required by practitioners in the energy sector, aligning with the timescale of
market bidding and system balancing \citep[see, e.g.,][]{elamin2018}; forecasts at coarser resolutions, such as daily forecasts, do not capture the intra-day dynamics that are crucial for these operational decisions.
The one-year-ahead horizon, on the other hand, is essential for energy companies to effectively plan their financial operations, particularly in terms of contract structuring and infrastructure investment \citep[see, e.g.,][]{hong2016}.

\titoletto{The dataset}
We conduct our analysis using two publicly available datasets to ensure the reproducibility of our results. One dataset contains total load data for the New England region, while the other contains net load data for the London area.
For both datasets, we examine two distinct forecasting approaches:
\textit{point} forecasting and \textit{probabilistic} forecasting. 
The former aims to generate predictions in the form of scalar values;
the latter in the form of probability distributions, thereby providing details about predictive uncertainty 
{\color{my_colour2}
and enabling risk-aware decision making
\citep[see, e.g.,][]{hong2019, makridakis2022m5}.
}

\titoletto{Section organisation}
In what follows, we first introduce and describe the two datasets, analysing their key features.
Then, we outline the forecasting method, providing a detailed description of the RNN($p$) setup -- specifically tailored to capture the temporal patterns typical of power consumption.
Finally, we present and discuss the forecasting results, focusing on both the predictive accuracy and the computational efficiency of the RNN($p$) models, and we compare the proposed approach against standard benchmarks.

\subsection{Datasets} \label{ssec:datasets}

\subsubsection{New England -- Load}
\titoletto{New England}
The first dataset we analyse refers to the \textit{total} load 
-- i.e.\ the total power consumption -- of the New England region 
\citep[cf.][]{hong2019}.
Load data are published by ISO New England (ISONE), the organisation responsible for overseeing the New England bulk electric power system.
The dataset contains hourly household electricity consumption (in MWh) from 2007 to 2014, along with recorded dry-bulb and dew-point temperatures (both in °F). For this study, we consider the time series of the aggregated load across the entire New England region.

\titoletto{Weather-dependent}
The forecasting task considered in this analysis consists in predicting electricity load given the realised weather conditions.
This setup is standard in the mid-term load forecasting literature, as it allows for an evaluation of model performance without embedding the meteorological forecasting errors \citep[see, e.g.,][]{goude2013}. By using realised weather data, we focus exclusively on modelling the relationship between weather conditions and electricity demand, ensuring a consistent basis for evaluating the predictive performance of different models 
\citep[see, e.g.,][]{behmiri2023}.

\smallskip
\titoletto{Dataset segmentation}
Following common practices in forecasting, the New England dataset is split into a training set (2007–2010), a validation set for hyperparameter selection (2011), and a test set (2012). In addition, the years 2013 and 2014 are reserved for further evaluating the robustness of the results. Descriptive statistics for total load and weather variables in the training set are provided in Table \ref{tab:descriptivestats1}.

\begin{table}[!h]
\vspace{0.1cm}
\centering
\begin{tabular}{L{6.cm} | R{1.85cm} R{1.85cm} R{1.85cm} R{1.85cm} R{1.85cm}}
	\midrule
	& Mean & Std.\ Dev. & Max & Median & Min \\
	\midrule
	Total Load [MWh] & 
		14713.91 & 2876.46 & 26705.00 & 14866.00 & 8894.00 \\
	Dry-bulb temperature [°F] & 
		49.66 & 18.57 & 96.62 & 50.37 & -4.37 \\
	Dew-point temperature [°F] & 
		37.80 & 19.27 & 74.25 & 39.12 & -19.50 \\
	\midrule
\end{tabular}
\caption{\small{Descriptive statistics for hourly load and temperatures in New England on the training set (2007-2010).}}
\label{tab:descriptivestats1}
\end{table}

\subsubsection{London -- Net Load}
\titoletto{London}
The second dataset we analyse refers to the \textit{net} load --
i.e.\ the total load minus the power produced from renewable sources -- for the London area from 2014 to 2019
\citep[cf.][]{browell2021}.
The dataset, published by the National Grid ESO, contains the net load,
together with the following weather variables: surface temperature, total precipitation,
cloud cover (low-level, mid-level, high-level), solar irradiance, and wind speed at 10m and 100m
\citep[see, e.g.,][for a detailed description of the dataset]{browell2021}. 
In our analysis, we consider aggregated hourly data.

As in the New England case, the London dataset is split into a training set (2014–2017), a validation set for hyperparameter selection (2018), and a test set (2019). Descriptive statistics for net load and weather variables in the training set are presented in Table \ref{tab:descriptivestats2}.

\begin{table}[!h]
\vspace{0.1cm}
\centering
\begin{tabular}{L{6.3cm} | R{1.65cm} R{1.65cm} R{1.65cm} R{1.65cm} R{1.65cm}}
	\midrule
	& Mean & Std Dev & Max & Median & Min \\
	\midrule
	Net Load [MWh] & 
		3310.48 & 755.66 & 5347.99 & 3355.55 & 1828.40 \\
	Temperature [K] & 
		284.02 & 5.73 & 305.14 & 284.10 & 267.22 \\
	Precipitation [mm/hour] & 
		0.27 & 0.94 & 24.44 & 0.00 & 0.00 \\
	Low Cloud Cover [\%] & 
		39.28 & 36.60 & 100.00 & 27.56 & 0.00 \\
	Mid Cloud Cover [\%] & 
		28.95 & 34.58 & 100.00 & 11.06 & 0.00 \\
	High Cloud Cover [\%] & 
		37.70 & 40.09 & 100.00 & 18.76 & 0.00 \\
	Solar Irradiance [W/m$^{2}$] & 
		126.30 & 191.74 & 876.59 & 9.41 & 0.00 \\
	Wind Speed 10m [m/s] & 
		4.07 & 1.98 & 15.05 & 3.75 & 0.29 \\
	Wind Speed 100m [m/s] & 
		6.85 & 2.98 & 23.20 & 6.58 & 0.38 \\
	\midrule
\end{tabular}

\caption{\small{Descriptive statistics for hourly net load and weather variables in London on the training set (2014-2017).}}
\label{tab:descriptivestats2}
\end{table}

\subsubsection{Seasonal and Autoregressive Features} \label{ssec:seasonalandautoregressive}

\begin{figure}[!h]
	\centering
	\includegraphics[width=0.99\linewidth, trim={0.2cm 0.2cm 0.2cm 0.2cm}, clip]{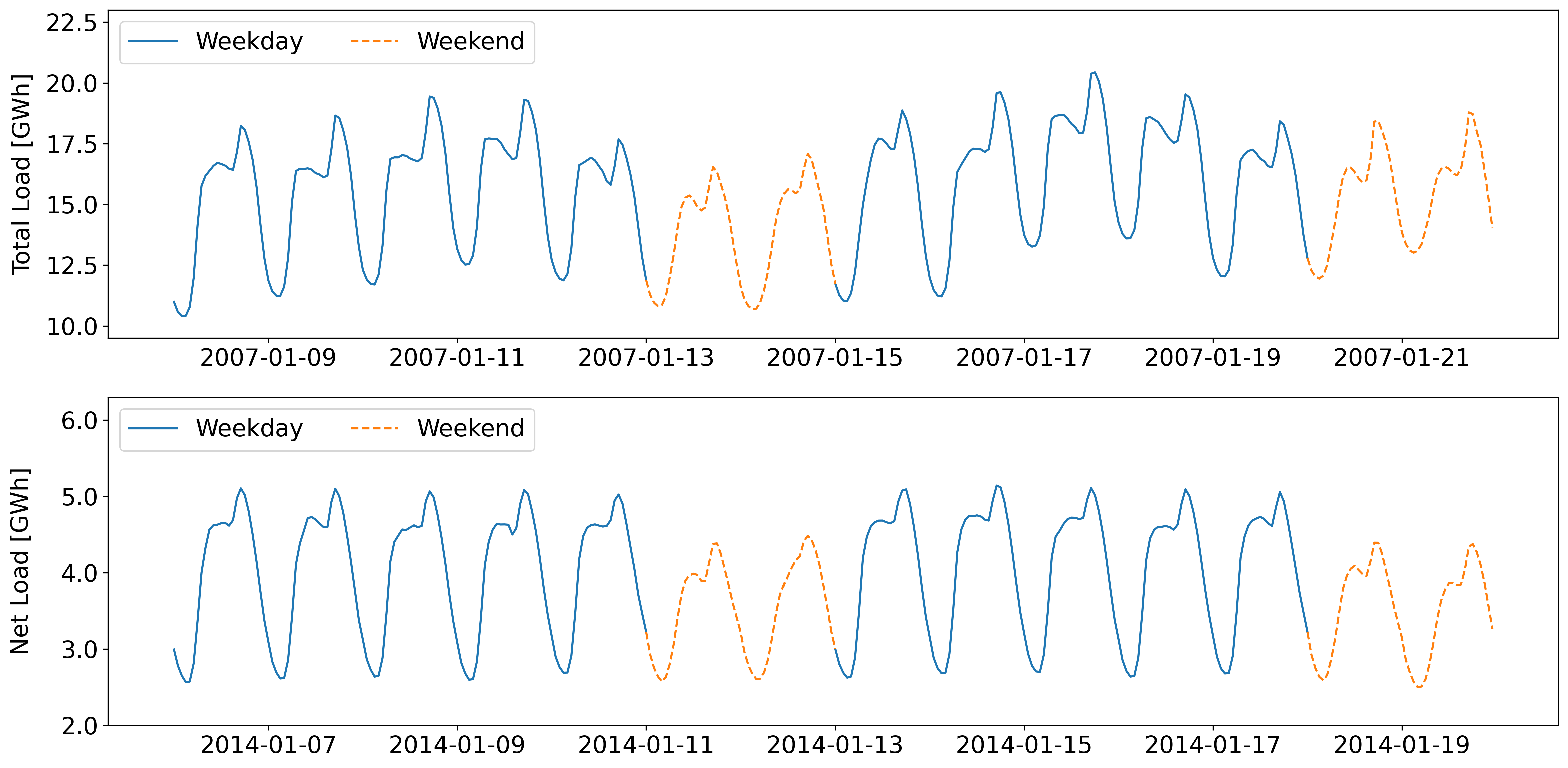}
\caption{\small{The first two full weeks of training data for the New England dataset (\textit{above}) and for the London dataset (\textit{below}). Both time series display evident intra-day seasonal patterns, characterised by peaks in the morning and late afternoon. Additionally, a weekly seasonality is observed, with noticeable load variations between weekdays (solid blue line) and weekends (dashed orange line).
}}
\label{fig:seasonality_pattens}
\end{figure}

\titoletto{Seasonality}
Seasonal dynamics in load time series are shaped by a combination of natural phenomena and human activities.
Figure \ref{fig:seasonality_pattens} illustrates the daily and weekly load patterns for the two datasets, considering the first two complete weeks of each training set.
Both time series show a highly distinctive behaviour: on the daily scale, we observe peaks in the morning and late afternoon; 
on the weekly scale, we observe a noticeable drop in load during the weekends.
Furthermore, the load curves for New England (\textit{above}) reveal a more pronounced difference between the two daily peaks compared to those for London (\textit{below}), where the profile is smoother during the central hours of the day, with a less noticeable dip between the two peaks.
This difference is likely attributable to the influence of solar energy production, which impacts net load but not total load.

\titoletto{Deseasonalisation}
As standard in the literature, the first step in our methodology is to remove the seasonal component from the load data, before processing the time series \citep[see, e.g.,][]{hong2016}.\footnote{
As widely noted in the forecasting literature, time series deseasonalisation can improve the performance of Neural Networks by allowing them to focus on the residual dynamics \citep[see, e.g.,][]{zhang2005}.
In Appendix~\ref{sec:appendixC}, we show that this holds also in our cases by comparing the results of RNN($p$) models with and without seasonal preprocessing.
}
The details of the deseasonalisation procedure are discussed in Section \ref{ssec:methodologyx}.

\titoletto{Autocorrelation}
Once removed the seasonality,
the residual time series is analysed to uncover any remaining temporal dependencies.
To do so, we measure the autocorrelation of the time series by means of the Partial AutoCorrelation Function (PACF), commonly used to determine the relevant autoregressive effects \citep[see, e.g.,][]{box2015}. This analysis is important for our approach because the identified temporal dependencies -- specifically the lag structure -- guide the selection of feedback connections in the RNN($p$) model.

\begin{figure}[!h]
	\centering
	\includegraphics[width=0.48\linewidth, trim={0.25cm 0.1cm 0.2cm 0.64cm}, clip]{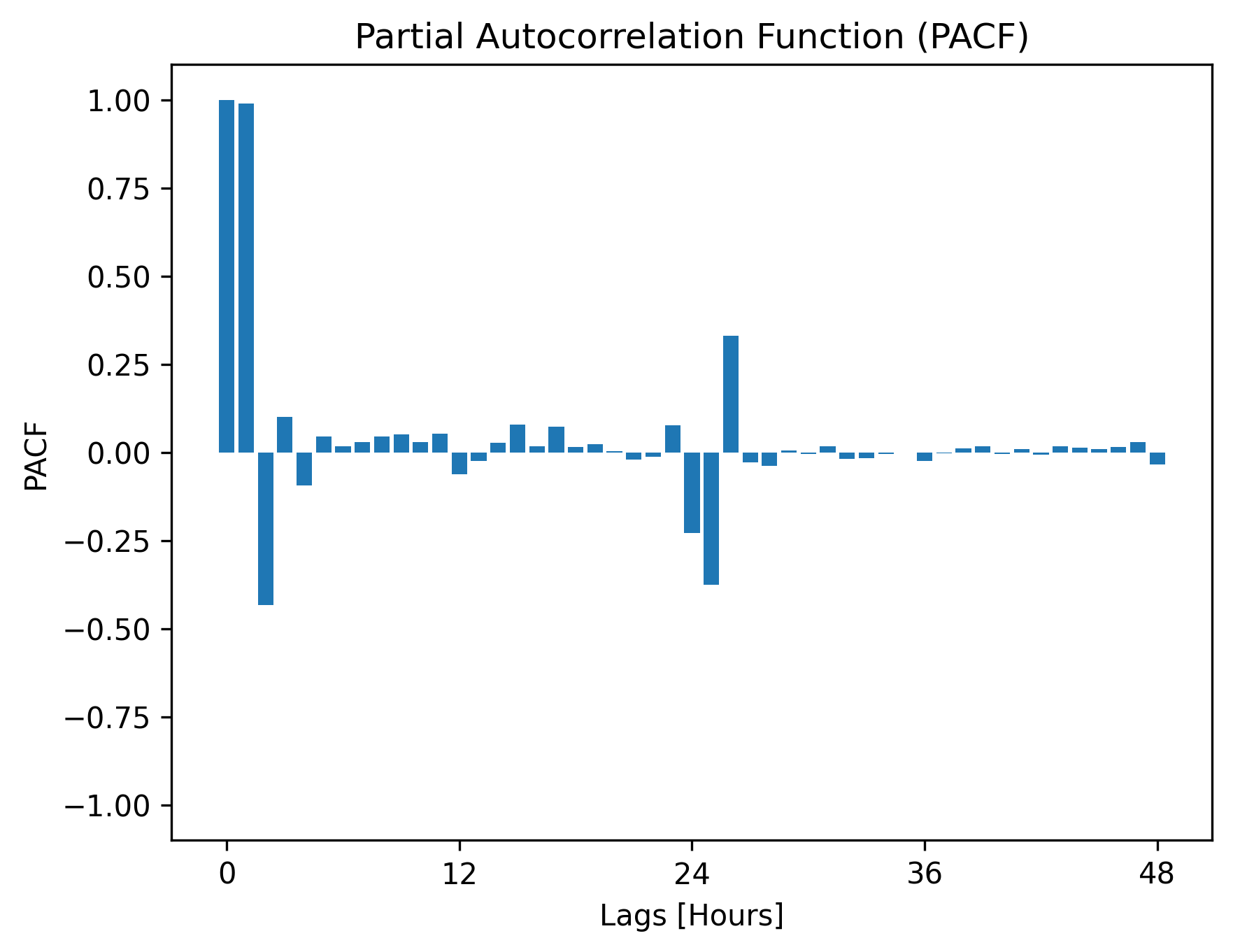}
	\hspace{0.2cm}
	\includegraphics[width=0.48\linewidth, trim={0.25cm 0.1cm 0.2cm 0.64cm}, clip]{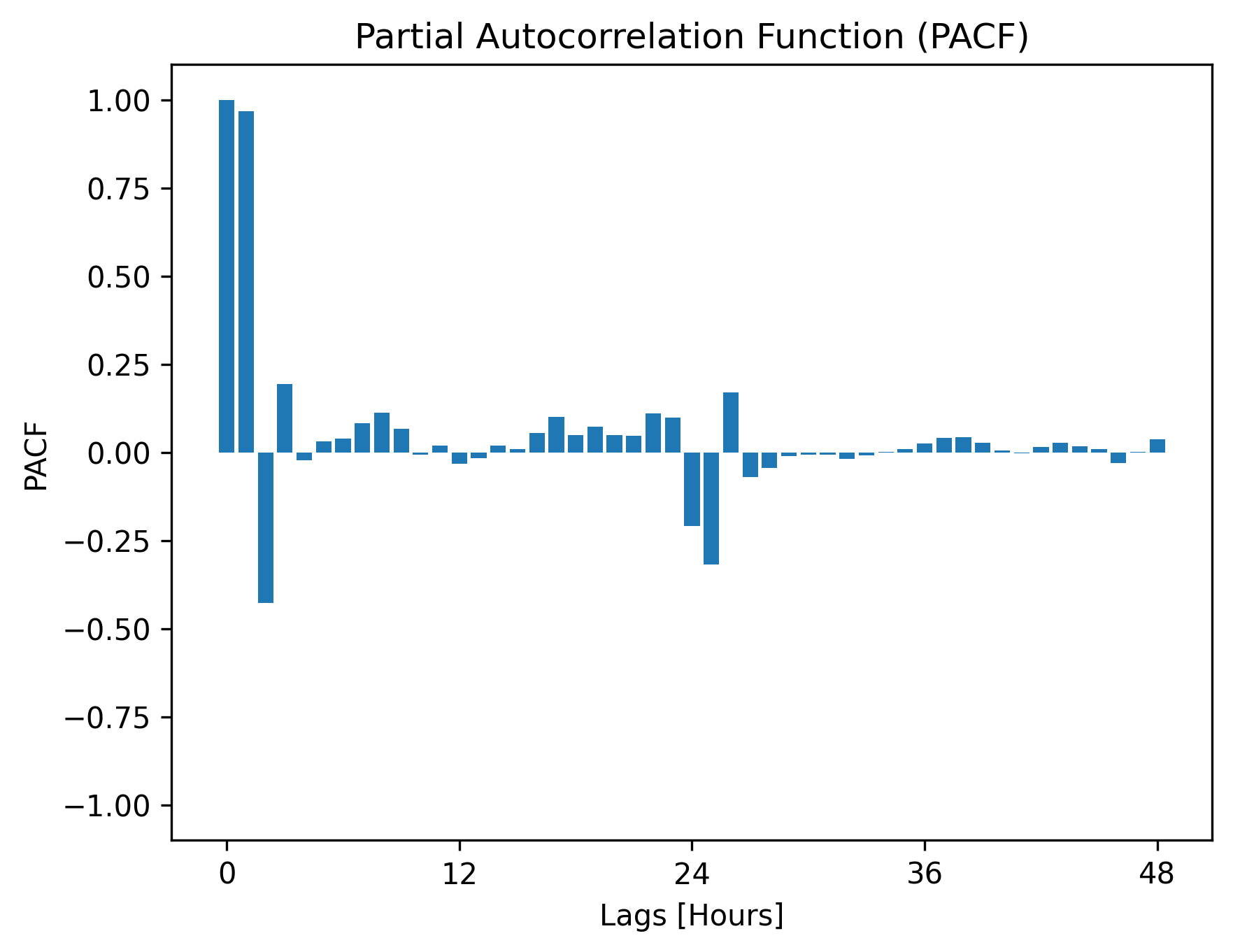}
	\caption{\small{Autocorrelation structure of the deseasonalised time series
	on the New England dataset (\textit{left}) and on the London dataset
	(\textit{right}). 
	PACF diagrams are obtained by analysing the two time series on the training set (2007-2010 and 2014-2017, respectively).
	The plots highlight the presence of a distinctive serial correlation, involving
	the first 2 lags (hourly autocorrelation) 
	and those around lag 24 (daily autocorrelation).
	}}
	\label{fig:PACF}
\end{figure}

For both datasets, the PACF on the training set is reported in Figure \ref{fig:PACF}.
As expected, it reveals two distinct autoregressive mechanisms, which involve respectively an hourly dependency
(the load in the two hours prior to time $t$)
and a daily dependency
(the load around the time $t{-}24$).
Furthermore, the Augmented Dickey-Fuller (ADF) test rejects the null hypothesis of a unit root at any relevant significance level, ensuring the stationarity of the two time series (with p-values lower than $10^{-15}$ in both cases).

\subsection{Forecasting method} \label{ssec:methodologyx}

The forecasting pipeline consists of the following steps: 
i) preprocessing of the load data, 
ii) training of the RNN($p$) on the deseasonalised data,
iii) construction of the final forecasts by combining the RNN($p$) forecasts with the seasonal baseline, and
iv) evaluation of the forecasts quality.

\smallskip
\titoletto{Step (i)}
i) Firstly, we preprocess the load data by removing the trend and seasonal component.
Following standard practice in electricity forecasting, we model the logarithm of the load \citep[see, e.g.,][]{benth2008, hong2016}.
This preprocessing step is performed using a linear model fitted on in-sample data, 
where trend and seasonality are removed separately for each of the twenty-four hours of the day. 
In line with recent literature, we include the following regressors: the first two Fourier terms for the day-of-the-year, three dummy variables to identify Saturdays, Sundays, and holidays, and a linear trend \citep[see, e.g.,][]{azzone2021, baviera2024daily}.

\smallskip
\titoletto{Step (ii)}
ii) Secondly, we process the time series using the RNN($p$). The network is trained using a moving window approach \citep[see, e.g.,][]{hewamalage2021}, where the data are split into overlapping sequences with length $\tau=49$ (corresponding to the current time step and the previous two days) to capture both the hourly and the daily dependencies observed in the data.

The inputs for the RNN($p$) are all available meteorological variables, along with calendar variables matching those used in the preprocessing: the first two Fourier terms for the day-of-the-year and the hour-of-the-day, and dummy variables to identify the day-of-the-week and holidays. To ensure consistent input ranges, all variables are min-max normalised based on in-sample data \citep[see, e.g.,][]{aggrawal2018}.

The output of the RNN($p$) depends on the considered forecasting approach. In the point forecasting case, the network outputs a single value for each time step -- the prediction of the log-load at time $t$ -- and is trained by minimising the Mean Squared Error (MSE), a standard loss function in load forecasting 
\citep[see, e.g.,][]{bianchi2017}.
In the probabilistic forecasting case, the log-load at time $t$ is modelled as a Gaussian random variable; the network outputs two values for each time step -- the mean $\hat{\mu}$ and standard deviation $\hat{\sigma}$ of the predicted distribution -- and is trained by minimising the Gaussian Negative Log-Likelihood (NLL).\footnote{
In the probabilistic forecasting case, the per-sample loss at each time step is 
$L = 0.5 \, \log(2 \pi \hat{\sigma}^2) + 0.5 \, {(y - \hat{\mu})^2}/ \hat{\sigma}^2$, where $y$ is the true log-load, while $\hat{\mu}$ and $\hat{\sigma}$ are the predicted distributional parameters.
%
%
In our implementation, for numerical stability, each predicted standard deviation $\hat{\sigma}$ is computed as 
$\hat{\sigma} = \abs{\tilde{\sigma}} + 10^{-9}$, where $\tilde{\sigma}$ is the raw network output; 
this ensures $\hat{\sigma} > 0$ while preserving the intended NLL optimisation. 
Alternative approaches, such as softplus or exponential activation functions, could also be used to ensure strict positivity.
}
{\color{my_colour2}
This modelling approach has proven to be very effective in load and price forecasting \citep[see, e.g.,][]{vossen2018, azzone2021, marcjasz2023distributional}.}\footnote{
The Gaussian distribution is here used for both analytical convenience and structural simplicity, as every forecast consists of a point estimate (mean) and a dispersion estimate (standard deviation). Nevertheless, for our datasets, diagnostic analyses indicate that the Gaussianity assumption is statistically appropriate (cf.\ Appendix~\ref{sec:appendixC}).
In any case, the RNN($p$) framework can easily accommodate other parametric distributions if different modelling assumptions are needed.
}

\smallskip
\titoletto{Step (iii)}
iii) Thirdly, we generate out-of-sample forecasts using the RNN($p$). These forecasts are then combined with the out-of-sample seasonal baseline to obtain the final prediction.
In the point forecasting setting, each final forecast is computed by adding the seasonal baseline to the RNN($p$) forecast.
In the probabilistic forecasting setting, each final forecast is a Gaussian distribution with mean equal to $\hat{\mu}$ plus the seasonal baseline, and standard deviation equal to $\hat{\sigma}$.
The obtained forecasts are then denormalised and exponentiated to return to the original scale.

We generate hourly predictions for the entire out-of-sample period (a full year ahead) in a single forecasting run.
As is standard in the mid-term forecasting literature 
\citep[see, e.g.,][]{goude2013}, the model is provided with true meteorological inputs but has no access to the true load values at any point in the out-of-sample period -- neither contemporaneous nor past. Instead, the model exclusively relies on its own prior load forecasts as autoregressive inputs for generating future forecasts over the entire year.

\smallskip
\titoletto{Step (iv)}
iv) Finally, we assess quantitatively the out-of-sample forecasts using standard performance metrics. In the point forecasting setting, we compute the Mean Absolute Percentage Error (MAPE) and the Root Mean Square Error (RMSE).
In the probabilistic forecasting setting, we also compute the Average Pinball Loss (APL) -- a measure of distributional accuracy obtained as the average of the Pinball Loss across percentiles -- and the NLL \citep[see, e.g.,][]{hong2019}.

A schematic of the full forecasting pipeline is provided in Figure \ref{fig:pipeline}.

\begin{figure*}[!h]
	\centering
\begin{tikzpicture}[
  node distance=0.6cm and 1.5cm,
  box/.style={draw=blue!50!black, fill=blue!5, rounded corners, minimum width=5cm, minimum height=0.9cm, align=center, font=\small},
  phase/.style={draw=gray!60, fill=gray!10, dashed, rounded corners, inner sep=0.3cm, fill opacity=0.2},
  arrow/.style={-{Latex[length=2mm]}, thick, blue!70!black},
  stepnum/.style={anchor=east, font=\small\bfseries, gray},
  phaselabel/.style={font=\small\bfseries, gray}
]

\node[box] (preproc) {Preprocessing};
\node[box, below=of preproc] (train) {\textbf{RNN($\mathbf{p}$)}};
\node[box, below=of train] (forecast) {Out-of-Sample Forecasts};
\node[box, below=of forecast] (eval) {Performance Assessment};

\node[box, left=of preproc] (inload) {In-Sample Load};
\node[box, left=of train] (inweather) {In-Sample Weather};
\node[box, left=of forecast] (outweather) {Out-of-Sample Weather};
\node[box, left=of eval] (outload) {Out-of-Sample Load};

\draw[arrow] (inload) -- (preproc);
\draw[arrow] (preproc) -- (train);
\draw[arrow] (inweather) -- (train);
\draw[arrow] (train) -- (forecast);
\draw[arrow] (outweather) -- (forecast);
\draw[arrow] (forecast) -- (eval);
\draw[arrow] (outload) -- (eval);

\node[anchor=east, font=\small] at ($(inload.west) + (-0.5,0)$) {i)};
\node[anchor=east, font=\small] at ($(inweather.west) + (-0.5,0)$) {ii)};
\node[anchor=east, font=\small] at ($(outweather.west) + (-0.5,0)$) {iii)};
\node[anchor=east, font=\small] at ($(outload.west) + (-0.5,0)$) {iv)};

\node[phase, fit={(inload) (inweather) (preproc) (train)}] (training) {};
\node[phase, fit={(outweather) (forecast)}] (forecasting) {};
\node[phase, fit={(outload) (eval)}] (evaluation) {};

\node[font=\small, anchor=west] at ($(training.east) + (0.1, 0)$) {\textit{Training}};
\node[font=\small, anchor=west] at ($(forecasting.east) + (0.1, 0)$) {\textit{Forecasting}};
\node[font=\small, anchor=west] at ($(evaluation.east) + (0.1, 0)$) {\textit{Evaluation}};

\node[font=\small, anchor=east] at ($(outload.west) + (-0.4, 0)$) {\phantom{\textit{Evaluation}}};

\end{tikzpicture}
	\caption{\small{
	Pipeline of the forecasting methodology.
	}}
	\label{fig:pipeline}
\end{figure*}

\bigskip

\titoletto{Hyperparameters (pt.1)}
As previously mentioned in Section \ref{ssec:datasets}, we adopt a standard training/validation/test approach in this analysis \citep[see, e.g.,][]{goodfellow2016}. The full forecasting pipeline in Figure \ref{fig:pipeline} is initially executed by fitting the model on the training set (which, for both datasets, corresponds to the first four years of data), while the validation set (the first year after the training set) is initially used as the out-of-sample data for hyperparameter selection. Specifically, to determine the optimal hyperparameters, we perform a grid search over the values reported in Table \ref{tab:hyperparams1}: the configuration that yields the best performance on the validation set 
--~measured in terms of MSE for point forecasting and NLL for probabilistic forecasting~-- 
is selected.

\titoletto{Hyperparameters (pt.2)}
Once the optimal hyperparameters are identified, the full forecasting pipeline is rerun. At this stage, the model is retrained on the combined training set and validation set (now treated as the in-sample data), and the forecasting performance is evaluated on the held-out test set, which serves as the final out-of-sample data.\footnote{All Neural Networks are trained using the Adam optimiser,
applying early stopping with a patience of 50 epochs, and a maximum training limit of 500 epochs. For computational efficiency, gradients are computed using AAD --  although, as previously underlined, all three algorithms discussed in Section \ref{sec:complex} are exact and yield identical numerical results.}

\begin{table}[!h]
\vspace{0.1cm}
\centering
\begin{tabular}{C{4.5cm} | C{7cm}}
    \midrule
    Hyperparameter & Values \\
    \midrule
    Activation Function & ReLU, Sigmoid \\
    Hidden Neurons & 
    	$ 5, 10, 15, 20$  \\
    Learning Rate & 
    	$0.0001, \, 0.0005, \, 0.001,  \, 0.005, \, 0.01$ \\
    Batch Size & 
    	$32, 64, 128$ \\
    \midrule
\end{tabular}
\caption{\small{Set of hyperparameters analysed in the grid search for both point and probabilistic forecasting.}}
\label{tab:hyperparams1}
\end{table}

\titoletto{Lag selection: general}
In addition to hyperparameter tuning, an important design choice in our setup 
is the selection of the multiple autoregressive lags (i.e.\ $p$), which are essential for capturing the multi-level dependencies in load data. 
As discussed earlier, this is a key feature of the RNN($p$) framework, as the use of
explicit feedbacks reduces the need for the model to relearn well-understood patterns, resulting in a simpler design.

\titoletto{Lag selection: the choice}
To assess the impact of different lag structures on forecasting performance, we compare three configurations for the RNN($p$) model: one with a single lag ($p=\{1\}$), one with two lags ($p=\{1,2\}$) and one with the full multi-lag structure ($p=\{1,2,24\}$). 
The selection of these lags is informed by the seasonal and autocorrelation analysis conducted in Section \ref{ssec:seasonalandautoregressive}, where clear daily periodicities and short-term dependencies are identified in the training data. In particular, the full model -- comprising lags 1,2, and 24 -- is designed to capture both the immediate past and the strong daily seasonality.\footnote{
An extensive comparison of alternative lag specifications is provided in Appendix \ref{sec:appendixC}.
}

\subsection{Results} \label{ssec:results}

In this section, we evaluate the forecasting accuracy and computational performance of the proposed RNN($p$) models. 
First, we present load forecasting results on the New England and London datasets, comparing the selected multi-lag architecture against standard benchmark models.
Then, we analyse the computational aspects of RNN($p$) training, reporting actual training times and comparing them with the theoretical complexity estimates presented in Section \ref{ssec:tcsc}.

To ensure the statistical robustness of the forecasting results, we train each model ten times for each architecture choice, using different random initialisations of the weights. We report the average values of the evaluation metrics, together with $\pm1$ Standard Error (SE).

\subsubsection{Forecasting Accuracy -- New England} \label{sec:forecasting_accuracy_new_england}

The results on the one-year test set (2012) reveal a striking improvement in forecasting accuracy, both in point and probabilistic case.

\smallskip
\titoletto{GEFCom: Point Forecasting}
The \textit{point forecasting} accuracy is summarised in Table \ref{tab:point_results}, which reports the  models' performance in terms of both MAPE and RMSE on the test set. The table illustrates the incremental improvements as we extend the autoregressive feedbacks to higher lags, specifically from $p=\{1\}$ to $p=\{1,2,24\}$. For this forecasting task, the optimal hyperparameters selected on the validation set are: sigmoid activation function, 10 hidden neurons, a learning rate of $0.001$, and a batch size of 32.\footnote{To isolate the effect of lag selection, the hyperparameter grid search is conducted for $p=\{1\}$, and the resulting configuration is applied uniformly to all considered multi-lag RNN($p$) models. This choice also reduces the overall computational burden of the validation procedure.}

While the addition of the second lag yields a modest improvement, the inclusion of both the second and twenty-fourth autoregressive feedbacks leads to a substantial gain in predictive accuracy.
As we observe, by simply enriching the temporal context of the model, the forecasting errors decrease by approximately 5\% compared to the single-lag variant, both in absolute and in relative terms. 
Furthermore, the achieved MAPE of 2.14\% is well below the value of 2.50\%, commonly regarded by practitioners as indicative of excellent mid-term load forecasts 
using realised weather conditions \citep[see, e.g.,][]{fan2012}.

\begin{table}[!h]
\centering
\begin{tabular}{ c | c c }
	\midrule
	Lags $p$ &
	MAPE [\%] & RMSE [MWh]  \\
	\midrule
	$\{1\}$ & 2.28 $\pm$ 0.02 & 468.87 $\pm$ 3.84 \\
	$\{1,2\}$ & 2.21 $\pm$ 0.02 & 459.41 $\pm$ 3.57 \\
	$\{1,2,24\}$ & \textbf{2.10 $\pm$ 0.01} & \textbf{441.14 $\pm$ 2.68} \\
	\midrule
\end{tabular}
\caption{\small{RNN($p$) \textit{point} forecasting results on the test set (2012) for
the New England dataset.}}
\label{tab:point_results}
\end{table}


\smallskip
\titoletto{GEFCom: Probabilistic Forecasting}
Even stronger results are obtained in the \textit{probabilistic forecasting} case.
Accuracy on the test set is assessed using MAPE and RMSE (computed for the means of the predicted distributions), and additionally with APL and NLL to evaluate the quality of the predicted distributions. For this forecasting task, the hyperparameters selected through the validation procedure are: sigmoid activation function, 10 hidden neurons, a learning rate of $0.001$, and a batch size of 64.

Table \ref{tab:prob_results} presents the performance on the test set for the three different RNN($p$) lag configurations.
Not only do multi-lag architectures achieve improved point forecasting accuracy
--~as already noted in Table \ref{tab:point_results}~-- 
but also the predicted densities are more appropriate to describe the distributions of the future consumption, as highlighted by the lower APL.
The full-lag model with $p = \{1,2,24\}$ reduces MAPE, RMSE and APL by nearly 10\% compared to the single-lag variant, indicating that the RNN is not only making better central forecasts but is also quantifying uncertainty more effectively. Additionally, the decrease in NLL confirms that the estimated distributions are better aligned with the data-generating process.
Finally, we note that both MAPE and RMSE remain comparable to those obtained with point forecasting.

The probabilistic forecasts for the first two full weeks of the test period are shown in Figure \ref{fig:probabilistic_forecasts_plot}.

\begin{table}[!h]
\centering
\begin{tabular}{ c | c c c c}
	\midrule
	Lags $p$ &
	MAPE [\%] & RMSE [MWh] & APL [MWh] & NLL \\
	\midrule
	$\{ 1 \}$ &
		2.26 $\pm$ 0.02 & 
		473.93 $\pm$ 3.11 & 
		122.10 $\pm$ 0.98 & 
		 7.47 $\pm$ 0.01 \\
	$\{ 1,2 \}$ &
		2.20 $\pm$ 0.03 & 
		463.19 $\pm$ 6.69 & 
		119.14 $\pm$ 1.75 &
		7.45 $\pm$ 0.01
		\\
	$\{ 1,2,24 \}$ &
		\textbf{2.08 $\pm$ 0.02} & 
		\textbf{436.55 $\pm$ 3.68} & 
		\textbf{111.62 $\pm$ 1.04} & 
		\textbf{7.39 $\pm$ 0.01}
		\\
	\midrule
\end{tabular}
\caption{\small{RNN($p$) \textit{probabilistic} forecasting results on the test set (2012) for
the New England dataset.}}
\label{tab:prob_results}
\end{table}

\titoletto{GEFCom: Comparison with Benchmarks (Probabilistic)}
To provide a comprehensive performance assessment, we compare RNN($p$) models against seven benchmarks, including both classical and state-of-the-art methods: 
a na\"ive baseline model, an ARX, a SARIMAX, an FNN, an LSTM, 
a LightGBM \citep[][]{ke2017lightgbm}, 
and a Transformer architecture
\citep[Single-DeT in][]{wang2022transformer}.
The na\"ive model generates forecasts based on historical load averages; all other benchmarks use the same set of exogenous variables as the RNN($p$). Complete implementation details for all benchmarks are provided in Appendix~\ref{sec:appendixC}.

The results of this comparison are reported in Table \ref{tab:comparison_benchmark_NE2012}.
Consistently across all metrics and years, the multi-lag RNN($\{1,2,24\}$) outperforms all other models.
Na\"ive forecasts are outperformed by a factor of three, confirming the limitations of overly simplistic models. 
The ARX and the SARIMAX models, despite their autoregressive structure, deliver only modest improvements: their forecasts are still worse by more than a factor of two compared to those by RNN($\{1,2,24\}$). This result emphasises the importance of capturing nonlinear dependencies to achieve accurate predictions.

Among the machine learning benchmarks, we observe that models that effectively capture temporal context deliver superior performance. 
LightGBM represents the first model to achieve competitive results, substantially improving accuracy over statistical techniques; however it is outperformed by all Neural Network-based models.
The FNN -- which captures nonlinearities but not temporal dependencies -- obtains the worst results among Neural Networks, with forecasts approximately 20\% worse than RNN($\{1,2,24\}$).
The single-lag RNN($1$), while improving upon the FNN by modelling sequence dynamics, still suffers from a restricted lag structure, resulting in forecasts roughly 10\% worse than RNN($\{1,2,24\}$).
Finally, the LSTM and the Transformer achieve performance similar to the RNN($\{1,2,24\}$), though slightly worse, with the added drawback of increased complexity and reduced interpretability.

\smallskip
Lastly, to further test the robustness of the obtained results, we conduct several additional analyses on the RNN($p$) models, which are detailed in Appendix~\ref{sec:appendixC}. These include evaluation on two further held-out test sets (years 2013 and 2014), alternative preprocessing choices, and a sensitivity analysis to lag specification.
Moreover, we also construct \textit{ensemble forecasts} by averaging, for each hour, the predictions across all 10 random-seed training runs \citep[see, e.g.,][]{richman2024}.
All robustness checks yield results consistent with the main findings, with the multi-lag RNN($\{1,2,24\}$) architecture maintaining its performance advantage across varied experimental conditions.

\begin{table}[!t]
\centering
\begin{tabular}{ l | c c c c}
\midrule
Model & MAPE [\%] & RMSE [MWh] & APL [MWh] & NLL \\
\midrule
Na\"ive & 6.66 & 1278.06 & 339.51 & 8.50 \\
ARX & 5.12 & 1050.61 & 281.29 & 8.33 \\
SARIMAX & 4.97 & 972.63 & 267.61 & 8.27 \\
LightGBM & 2.92 $\pm$ 0.01 & 546.51 $\pm$ 0.96 & 145.75 $\pm$ 0.32 & 7.68 $\pm$ 0.01 \\
FNN & 2.65 $\pm$ 0.04 & 564.17 $\pm$ 8.65 & 141.75 $\pm$ 2.06 & 7.62 $\pm$ 0.01 \\
RNN($1$) & 2.26 $\pm$ 0.02 & 473.93 $\pm$ 3.11 & 122.10 $\pm$ 0.98 & 7.47 $\pm$ 0.01 \\
RNN($\{1,2,24\}$) & $\mathbf{2.08 \pm 0.02}$ & $\mathbf{436.55 \pm 3.68}$ & $\mathbf{111.62 \pm 1.04}$ & $\mathbf{7.39 \pm 0.01}$ \\
LSTM & 2.18 $\pm$ 0.04 & 467.66 $\pm$ 10.78 & 118.76 $\pm$ 2.45 & 7.45 $\pm$ 0.02 \\
Transformer & 2.19 $\pm$ 0.04 & 463.25 $\pm$ 5.69 & 119.06 $\pm$ 2.04 & 7.50 $\pm$ 0.02  \\
\midrule
\end{tabular}

\caption{Comparison of RNN($p$) with benchmarks on the New England dataset: results of probabilistic forecasting on the test set (2012).
The best forecasts are highlighted in bold.}
\label{tab:comparison_benchmark_NE2012}
\end{table}

\subsubsection{Forecasting Accuracy -- London}
\titoletto{London: Point Forecasting}
We perform an analogous analysis on the London dataset referring to net load.
The results on the one-year test set (2019) are equally impressive, confirming the efficacy of the proposed forecasting approach.

\smallskip
Table \ref{tab:point_results_GB} reports the \textit{point forecasting} performance of the RNN($p$) models under the same feedback configurations used in the New England case. 
For this task, the optimal hyperparameters selected by the validation procedure are: sigmoid activation function, 15 hidden neurons, learning rate equal to $0.001$, and batch size equal to 32.
 
Analogously to the previous dataset,
the inclusion of additional autoregressive feedbacks leads to a clear improvement in point forecasting performance. The RNN($p$) configuration incorporating all three lags ($p = \{1, 2, 24\}$) consistently outperforms the simpler alternatives, yielding lower MAPE and RMSE scores.
Overall, this setup obtains a forecast accuracy improvement of approximatively 7\%, both in absolute and relative terms.

\begin{table}[!ht]
\centering
\begin{tabular}{ c | c c }
	\midrule
	Lags $p$ &
	MAPE [\%] & RMSE [MWh]  \\
	\midrule
	$\{1\}$ & 3.15 $\pm$ 0.07 & 136.34 $\pm$ 2.81 \\
	$\{1,2\}$ & 3.12 $\pm$ 0.07 & 133.83 $\pm$ 3.07 \\
	$\{1,2,24\}$ & \textbf{3.01 $\pm$ 0.05} & \textbf{126.87 $\pm$ 1.56} \\
	\midrule
\end{tabular}
\caption{\small{RNN($p$) \textit{point} forecasting results on the test set (2019)
for the London dataset.}}
\label{tab:point_results_GB}
\end{table}

\titoletto{London: Probabilistic Forecasting}
\textit{Probabilistic forecasting} results for the London dataset are shown in Table \ref{tab:prob_results_GB}, which reports model performance in terms of MAPE, RMSE, APL, and NLL.
For this task, the optimal hyperparameters selected by the validation procedure are: sigmoid activation function, 10 hidden neurons, learning rate equal to $0.005$, and batch size equal to 64.

Compared to the baseline single-lag configuration, the full model ($p = \{1, 2, 24\}$) achieves a 9\% reduction in RMSE and of 7\% in MAPE, indicating probabilistic forecasts that are better centred around the true values. Additionally, we observe a decrease in both APL and NLL, suggesting a better overall fit of the predicted distributions.

\begin{figure}[!t]
	\centering
	\includegraphics[width=0.99\linewidth, trim={0.2cm 0.2cm 0.2cm 0.2cm}, clip]{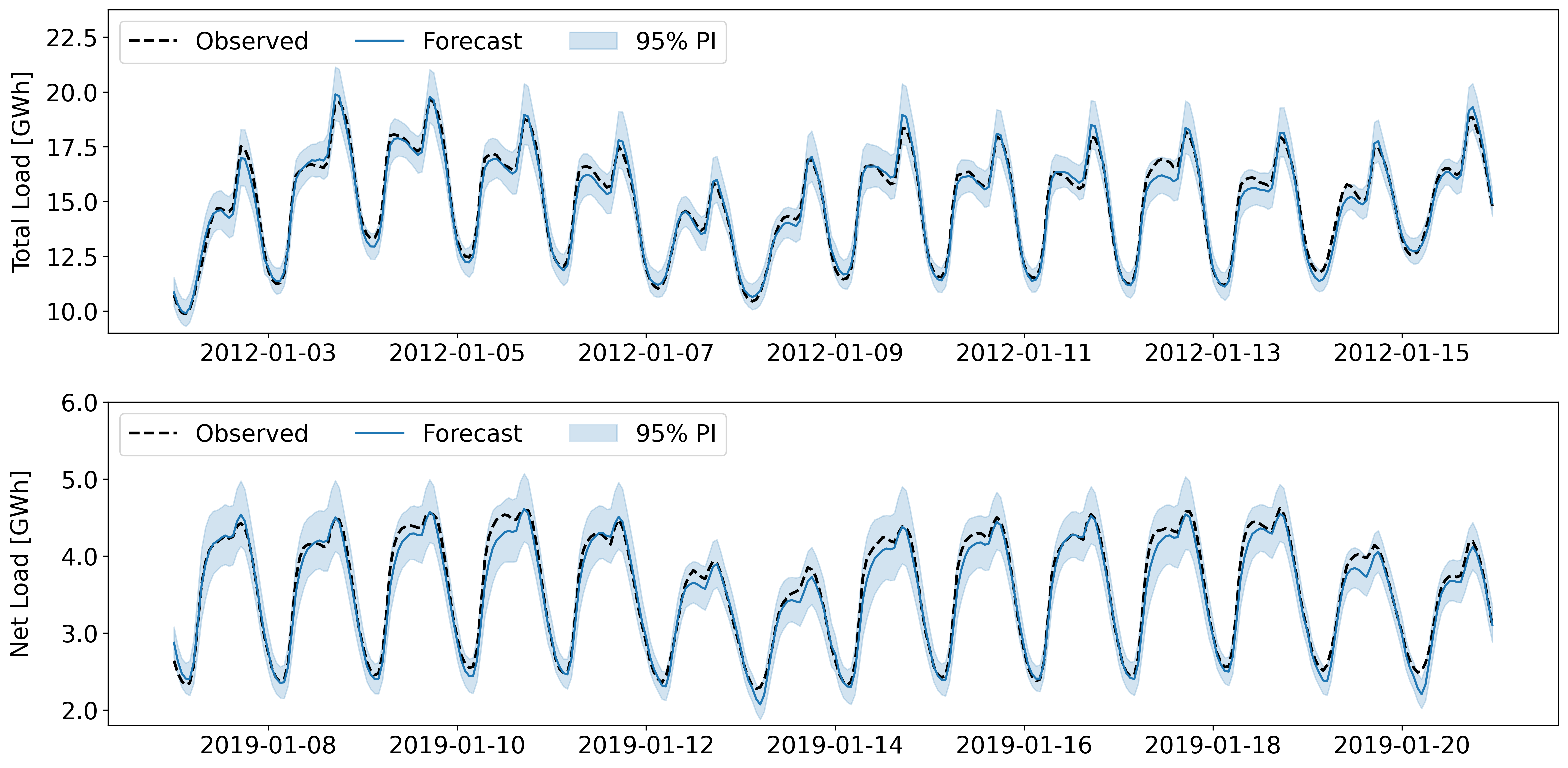}
\caption{\small{Probabilistic forecasts for the first two full weeks of the test period, 
generated by the RNN($\{1, 2, 24\}$), for the New England dataset (\textit{above}) and for the London dataset (\textit{below}).
The dashed black line indicates the observed load, the blue line the predicted mean, and the shaded area the 95\% Prediction Intervals (PIs). The mean forecasts closely follow the intra-daily load patterns, and the observed values fall within the 95\% PIs in nearly all cases.
}}
\label{fig:probabilistic_forecasts_plot}
\end{figure}

The probabilistic forecasts for the first two full weeks of the test period are shown in Figure \ref{fig:probabilistic_forecasts_plot}.

\begin{table}[!ht]
\centering
\begin{tabular}{ c | c c c c}
	\midrule
	Lags $p$ &
	MAPE [\%] & RMSE [MWh]  & APL [MWh] & NLL \\
	\midrule
	$\{1\}$ & 3.15 $\pm$ 0.06 & 136.80 $\pm$ 3.69 & 36.44 $\pm$ 1.57 & 6.21 $\pm$ 0.02\\
	$\{1,2\}$ & 3.11 $\pm$ 0.05 & 133.08 $\pm$ 2.11 & 35.71 $\pm$ 0.56  & 6.20 $\pm$ 0.02 \\
	$\{1,2,24\}$ & \textbf{2.97 $\pm$ 0.03} & \textbf{125.18 $\pm$ 1.30} & 
	\textbf{33.64 $\pm$ 0.39} & \textbf{6.17 $\pm$ 0.01} \\
	\midrule
\end{tabular}
\caption{\small{RNN($p$) \textit{probabilistic} forecasting results on the test set (2019)
for the London dataset.}}
\label{tab:prob_results_GB}
\end{table}

Finally, in Table \ref{tab:prob_results_GB_benchmark}, we compare the predictive performance of RNN($p$) models and those obtained by the same seven benchmark models described in Section  \ref{sec:forecasting_accuracy_new_england}.
The results are consistent with those obtained with the New England dataset.
The na\"ive benchmark delivers the poorest performance and is outperformed by a significant margin by all models. 
The RNN($\{1,2,24\}$) obtains forecasts that are consistently twice as accurate as those produced by the linear ARX and SARIMAX models, demonstrating the advantage of modelling nonlinear dependencies. 
The models which are not designed to handle sequential dependencies --~namely LightGBM and FNN~-- obtain intermediate results, confirming the importance of capturing temporal dynamics for this forecasting task.
Finally, the RNN($\{1,2,24\}$) also surpasses both the LSTM and Transformer, offering not only superior forecasting performance but also better interpretability in comparison to these more complex architectures.

\begin{table}[!t]
\centering
\begin{tabular}{ l | c c c c}
\midrule
Model & MAPE [\%] & RMSE [MWh] & APL [MWh] & NLL \\
\midrule
Na\"ive 
& 6.77 
& 256.42 
& 72.31 
& 6.99 \\
ARX 
& 5.28
& 211.98
& 58.44
& 6.78 \\
SARIMAX & 5.56 & 218.48 & 66.01 & 6.96 \\
LightGBM & 3.43 $\pm$ 0.01 & 132.45 $\pm$ 0.24 & 36.67 $\pm$ 0.08 & 6.25 $\pm$ 0.01 \\
FNN 
& 3.24 $\pm$ 0.06 
& 145.91 $\pm$ 2.98 
& 37.69 $\pm$ 0.83
& 6.27 $\pm$ 0.03 \\
RNN($1$) 
& 3.15 $\pm$ 0.06 
& 136.80 $\pm$ 3.69 
& 36.44 $\pm$ 1.57 
& 6.21 $\pm$ 0.02 \\
RNN($\{1,2,24\}$) 
& \textbf{2.97 $\pm$ 0.03} 
& \textbf{125.18 $\pm$ 1.30} 
& \textbf{33.64 $\pm$ 0.39} 
& \textbf{6.17 $\pm$ 0.01} \\
LSTM 
& 3.13 $\pm$ 0.04
& 138.60 $\pm$ 2.25 
& 36.14 $\pm$ 0.56
& 6.24 $\pm$ 0.02 \\
Transformer
& 3.12 $\pm$ 0.04
& 140.12 $\pm$ 2.26
& 35.98 $\pm$ 0.55
& 6.22 $\pm$ 0.02 \\
\midrule
\end{tabular}
\caption{\small{Comparison of RNN($p$) with benchmarks on the London dataset: results of probabilistic forecasting on the test set (2019).
The best forecasts are highlighted in bold.}}
\label{tab:prob_results_GB_benchmark}
\end{table}

\bigskip
The improvements in forecasting accuracy reported above have direct economic implications. 
To provide a concrete, albeit partial, estimate of their economic value, we consider a back-of-the-envelope calculation based only on imbalance settlement costs 
--~the costs resulting from differences between forecasted and actual load, which the system operator must cover in real time.
Following \citet[][]{matijavs2011mean}, a reduction $\Delta\text{MAPE}$ in forecast error translates into approximate annual savings of
\begin{equation*}
\text{Savings} \simeq \frac{\Delta \text{MAPE}}{100} \times \overline{\Lambda} \times \overline{P}^{\text{imb}} \times 8760 \,,
\end{equation*}
where $\overline{\Lambda}$ is the average hourly system net load (in MWh), $\overline{P}^{\text{imb}}$ the average imbalance price (in \pounds\,/\,MWh), and $8760$ the number of hours in a year.

{ \color{my_colour2}
For the London dataset, using an average imbalance price of 50 \pounds\,/\,MWh 
--~consistent with 2019 UK values \citep[cf.][Fig.~3, p.~7]{chen2024examining}~-- 
and the average net load reported in Table~\ref{tab:descriptivestats2}, this corresponds to roughly \pounds 15~million per percentage point reduction in MAPE every year.
This estimate indicates that transitioning from standard statistical models (ARX, SARIMAX)
--~still widely used in the power sector~-- to the proposed RNN($p$) framework can yield annual savings of several tens of millions of pounds for London alone.

\smallskip
The imbalance cost estimate above is based solely on point forecast accuracy, and therefore understates the full economic value of the improvements achieved by the RNN($p$) models in probabilistic forecasting.
The availability of accurate distributional forecasts
$(\hat{\mu}_t, \hat{\sigma}_t)$ enables a broader set of financial decision-making tools. 
In particular, the predicted standard deviations $\hat{\sigma}_t$ provide a time-varying measure of uncertainty that can be incorporated into risk-adjusted bidding strategies in day-ahead and intraday electricity markets \citep[see, e.g.,][]{conejo2010decision}, and the predicted distributions can serve as direct inputs for pricing load-following derivatives and constructing forward price curves \citep[see, e.g.,][]{howison2009stochastic, Nowotarski2018}.
These applications rely heavily on the quality of the distributional forecasts: consequently, the improvements in APL and NLL reported in Tables~\ref{tab:comparison_benchmark_NE2012} and~\ref{tab:prob_results_GB_benchmark} 
carry economic relevance beyond what MAPE and RMSE alone convey.
}

Finally, in addition to the financial benefits, improved forecasts also provide substantial environmental benefits by reducing reliance on carbon-intensive generation and enabling more efficient integration of renewable energy sources.

\subsubsection{Computational Comparison}

\titoletto{Computational Comparison}
To conclude the experimental analysis, which so far has focused on forecasting accuracy of RNN($p$), we conduct a final evaluation of their computational efficiency.\footnote{
Models are implemented in C++20 and trained on a machine with Apple M1 Chip, 8 GB RAM, macOS Sequoia 15.6.1.
} 
Specifically, we empirically test the three learning algorithms presented in Section \ref{sec:complex}, validating the complexity estimates derived in Proposition \ref{prop:complexities}.

First, we compare the observed training time of the algorithms for different values of the sequence length $\tau$.
Figure \ref{fig:plot_tau} reports the training time, expressed in seconds per epoch, required by a RNN($2$) with $10$ hidden neurons ($h=10$); we consider the whole New England training set ($x=17$) in the probabilistic forecasting case ($y=2$).
As expected, the computational time required by BPTT grows exponentially in 
$\tau$, entailing the practical impossibility of using this algorithm for training RNN($p$) when $p>1$.
Conversely, the time complexity associated with AAD and RTRL grows linearly with
$\tau$, in accordance with the complexity estimates proven in Proposition \ref{prop:complexities}.
\begin{figure}[!ht]
	\centering
	\resizebox{0.5\linewidth}{!}{
\begin{tikzpicture}
  \begin{axis}[
 	xmin= 2, xmax=26,
 	ymin=-0.5, ymax=12.5,
  	xtick={0,4,...,24},
  	xticklabels={,4,8,12,16,20,24},
	xlabel={$\tau$},
	ylabel={Seconds per epoch},
	width=10cm,
	height=7cm
  ]
    \addplot+[green, mark=*, mark options={solid, fill=green}, mark size=1.2pt, line width=0.75pt] coordinates {
      ( 3, 0.13)
      ( 4, 0.16)
      ( 5, 0.19)
	  ( 6, 0.22)
	  ( 7, 0.25)
	  ( 8, 0.28)
	  ( 9, 0.31)
	  (10, 0.34)
	  (11, 0.37)
	  (12, 0.40)
	  (13, 0.42)
	  (14, 0.45)
	  (15, 0.48)
	  (16, 0.51)
	  (17, 0.54)
	  (18, 0.57)
	  (19, 0.60)
	  (20, 0.63)
	  (21, 0.66)
	  (22, 0.69)
	  (23, 0.72)
	  (24, 0.75)
    };
    \addlegendentry{AAD}
    \addplot+[red, mark=square*, mark options={solid, fill=red}, mark size=1pt, line width=0.75pt, densely dotted] coordinates {
      ( 3, 0.24)
      ( 4, 0.30)
      ( 5, 0.35)
	  ( 6, 0.41)
	  ( 7, 0.47)
	  ( 8, 0.52)
	  ( 9, 0.59)
	  (10, 0.64)
	  (11, 0.69)
	  (12, 0.76)
	  (13, 0.81)
	  (14, 0.87)
	  (15, 0.93)
	  (16, 0.98)
	  (17, 1.04)
	  (18, 1.10)
	  (19, 1.16)
	  (20, 1.21)
	  (21, 1.27)
	  (22, 1.33)
	  (23, 1.40)
	  (24, 1.47)
    };
    \addlegendentry{RTRL}
    \addplot+[blue, mark=triangle*, mark options={solid, fill=blue}, mark size=2pt, line width=0.75pt, densely dashed] coordinates {
      ( 3, 0.15)
      ( 4, 0.21)
      ( 5, 0.30)
	  ( 6, 0.45)
	  ( 7, 0.68)
	  ( 8, 1.11)
	  ( 9, 1.68)
	  (10, 2.87)
	  (11, 4.49)
	  (12, 7.14)
	  (13, 11.31)
	  };
   \addlegendentry{BPTT}
  \end{axis}
\end{tikzpicture}
	}
	\caption{\small{
	Observed training time (in seconds per epoch) against 
	sequence length $\tau$ for the three algorithms. 
	We consider an RNN for probabilistic forecasting ($y=2$) with 10 hidden
	neurons and two autoregressive feedbacks ($p=2$).
	AAD and BPTT have similar behaviour when $\tau$ is very small 
	$(\tau \leq 4$): in this case, they are both faster than RTRL.
	However, when we process long sequences, the time-per-epoch grows
	exponentially for BPTT, as a consequence of the $p$-bonacci explosion;
	instead, the growth is approximatively linear for AAD and RTRL.
	}}
	\label{fig:plot_tau}
\end{figure}

\smallskip
\titoletto{AAD vs RTRL}
As a second test, we conduct a comparison on the training time required by AAD and RTRL.\footnote{The time-per-epoch in the case of BPTT is not reported for the infeasibility of the algorithm due to its exponential complexity (except in the case $p=1$, when BPTT coincides with AAD).} 
Table \ref{tab:speed_results} reports the measured average time-per-epoch required for training a RNN($p$), considering the three different lag configurations $p=\{1\}$, $p=\{1,2\}$ and $p=\{1,2,24\}$.
We test the computational time required to process one epoch of data, as a function of the number of hidden neurons and of recurrent connections, using again the New England training set in the probabilistic setting.\footnote{Analogous empirical results are found in the case of point forecasting. These results are not reported in the paper, and are available upon request.}
As expected, AAD consistently proves to be more efficient than RTRL, in particular as the number of hidden neurons $h$ increases.
We observe that the former algorithm is faster than the latter by a factor (called \textit{speedup factor} in the table) which is very close to the ratio $p y$, computed analytically in Section \ref{sec:complex}. 
Finally, we observe that increasing the number of feedbacks $p$ results in a clear linear increase in computational time for RTRL, consistent with its theoretical time complexity. In contrast, the training time required by AAD remains nearly constant as $p$ increases, highlighting its ability to efficiently handle multi-lags models.

\begin{table*}[!ht]
\vspace{0.1cm}
\centering
\begin{tabular}{c|cccc|cccc|c}
\toprule
\multirow{2}{*}{Lags $p$} 
& \multicolumn{4}{c|}{\textbf{AAD} (sec/epoch)} 
& \multicolumn{4}{c|}{\textbf{RTRL} (sec/epoch)} 
& \multirow[c]{2}{*}{\shortstack{Speedup \\ factor}} \\
\cmidrule{2-9}
& $h{=}5$ & $h{=}10$ & $h{=}15$ & $h{=}20$ 
& $h{=}5$ & $h{=}10$ & $h{=}15$ & $h{=}20$ & \\
\midrule
$\{1\}$         & 0.56 & 1.08 & 1.65 & 2.11  & 1.50 & 2.79 & 4.16 & 5.19  & 2 \\
$\{1,2\}$       & 0.68 & 1.20 & 1.81 & 2.40  & 2.71 & 5.31 & 7.92 & 10.51 & 4 \\
$\{1,2,24\}$    & 0.72 & 1.27 & 1.95 & 2.57  & 4.32 & 8.96 & 13.13 & 16.80 & 6 \\
\bottomrule
\end{tabular}
\caption{\small{
	Time-per-epoch required by learning algorithms as a function of the number of hidden neurons $h$
	when training in the probabilistic framework ($\tau=49$) on the New England dataset.
	We observe that for AAD the computational time does not depend on the order $p$ of the RNN. Moreover, the speedup factor of AAD with respect to RTRL is approximately equal to the theoretical one, indicated in the last column. 
	}}
\label{tab:speed_results}
\end{table*}

\section{Conclusions} \label{sec:conclusions}

In this paper, we have introduced RNN($p$) models, a family of Recurrent Neural Networks with Jordan feedbacks that represent the natural nonlinear extension of the well-known ARX($p$) models. The RNN($p$) architecture retains the high interpretability of ARX models while demonstrating improved forecasting accuracy.

The main contributions of this work are threefold. First, we have demonstrated that RNN($p$) models offer a powerful approach for capturing sequential dependencies in time series displaying an inherent seasonality, with a clear advantage in interpretability compared to complex black-box models. 

Second, we have analysed three different learning algorithms for training RNN($p$) models: RTRL, BPTT and AAD.
The simplicity of the considered architecture has allowed us to compute explicitly the time complexity and the space complexity associated to the execution of these algorithms.
We have explained the inefficiency of BPTT due to the exponential growth in time complexity with sequence length. 
Furthermore, we have quantified the improvement of AAD, deducing analytically the leading order of time complexity and space complexity.

Finally, we have applied the proposed RNN($p$) models to power consumption forecasting, considering two public datasets, one referring to the total load in New England, the other referring to the net load in London. 
We have shown that these models can effectively handle long-time dependencies, achieving high-precision forecasts on an hourly scale, both in terms of point and probabilistic forecasts. 

Our research has demonstrated that RNN($p$) models are well-suited for capturing the nonlinear dynamics found in real-world time series forecasting, emphasising their practical significance in energy markets and other fintech applications, where accurate and interpretable forecasting plays a critical role in decision-making processes.
Building on this foundation, future developments will focus on a systematic study of numerical stability and gradient dynamics, aiming to further enhance the robustness and scalability of RNN($p$) models for broader forecasting applications.

\section*{Acknowledgments}
\noindent 
We are grateful to 
C.\ Alasseur, M.\ Azzone, 
J.\ Blomvall, G.\ Consigli, 
A.\ Falcetta, Y.\ Goude,
O.\ Hammarlid and M.\ Roveri
for helpful suggestions. We would also like to thank all participants to the conferences:
``INREC14'' (Essen),
``WPI-Workshop: Stochastics, Statistics \& Machine Learning'' (Vienna),
``ECSO-CMS 2024 Conference'' (Stockholm), and ``International Fintech Research Conference'' (Perugia).
Finally, a special thanks goes to the members of the team R33 at EDF R\&D (Paris) as well as to all attendees to the EDF seminar ``RNN($p$) for Power Consumption Forecasting'' for stimulating questions and comments.

\newpage
\clearpage
 
\bibliography{my_bibliography.bib}

\appendix

\newpage
\section{Proofs} \label{sec:appendixA}

\setcounter{equation}{0}
\setcounter{theorem}{0}
\renewcommand{\theequation}{\thesection.\arabic{equation}}
\renewcommand{\thetheorem}{\thesection.\arabic{theorem}}

\bigskip
\paragraph{Proof of Proposition \ref{prop:complexities}}

In the following, we deduce the time and space complexity for the three learning algorithms separately 
-- part (a) for AAD, part (b) for RTRL, part (c) for BPTT -- 
based on their implementations provided in Appendix \ref{sec:appendixALGO}.

\bigskip \noindent
(a) \underline{AAD}.
The time complexity of AAD is determined by the operations at lines 3, 4 and 5 of Algorithm 1 in Appendix \ref{sec:appendixALGO}.
At each iteration of the \textit{for} loop, 
the involved time complexities are as follows:
\begin{itemize} \itemsep0em
\item[(i)] The execution of line 3 has time complexity 
$\mathcal{O}(p\, h\, y)$. 
Indeed, each of the $p$ vectors in the summation is the result of three matrix-vector multiplications.
First, we multiply the vector
$\dvinline{L}{\vvh{y}{t+k}} \in \mathbb{R}^{1 \times y}$ 
and the matrix
$V \in \mathbb{R}^{y \times h}$, 
with a complexity of $\mathcal{O}(h\, y)$.
Second, we multiply the obtained vector, 
which belongs to $\mathbb{R}^{1 \times h}$, 
and the diagonal matrix 
$A^{(t+k)} \in \mathbb{R}^{h \times h}$,
with a complexity of $\mathcal{O}(h)$.
Third, we multiply the obtained vector,
which belongs to $\mathbb{R}^{1 \times h}$,  
and the matrix $W_k \in \mathbb{R}^{h \times y}$, 
with a complexity of $\mathcal{O}(h\, y)$.
\item[(ii)] The execution of line 4 has time complexity 
$\mathcal{O}(h\, \vartheta)$.
Indeed, similarly to step (i), we perform three matrix-vector multiplications, with complexities $\mathcal{O}(h\, y)$, $\mathcal{O}(h\, y)$ and $\mathcal{O}(h\, \vartheta)$.
\item[(iii)] The execution of line 5 involves only a matrix-vector multiplication, and has time complexity $\mathcal{O}(y\, \varphi)$.
\end{itemize}
We observe that step (i) has a time complexity of 
$\mathcal{O}(p\, h\, y)$, where $p,\ h\, y < \vartheta$ by definition \eqref{eq:wtotalweights}.
Therefore, the leading terms are those relative to steps (ii) and (iii). 
As already discussed when describing the RNN($p$) architecture in Section   \ref{sec:model_spec}, $y$ is always less than $h$; thus, we can approximate the complexity of (iii) from above as $\mathcal{O}(h\, \varphi)$, obtaining an overall (upper bound for the) time complexity of $\mathcal{O}(h\, w)$.

\smallskip
In terms of space complexity, AAD is a backward-operating algorithm that requires access to all the previous inputs $\vvv{x}{t}$,
hidden states $\vvv{h}{t}$, and outputs $\vvh{y}{t}$. 
For a sequence of length $\tau$, this yields a space complexity of 
$\mathcal{O}(\tau\, (x+h+y) )$.

\bigskip \noindent
(b) \underline{RTRL}.
The time complexity of RTRL is determined by the computation of the total derivatives
at lines 2 and 3.
We perform the matrices multiplication from left to right.

\noindent
Let us first consider the first equation.
We observe that:
\begin{itemize} \itemsep0em
\item[(i)]
The computation of the matrix $VA^{(t)}$ 
has time complexity $\mathcal{O}(h\, y)$.
Indeed, the matrix $V$ has size  $y \times h$, while the matrix 
$A^{(t)}$ has size $h \times h$ and is diagonal because the activation function $\actf(\boldsymbol{\cdot})$ acts element-wise. 
\item[(ii)]
The computation of the matrix in brackets has time complexity
$\mathcal{O}(p\, h\, y\, \vartheta)$.
Indeed, the multiplication of each of the $p$ full matrices
$W_i \,\, \dvinline{\vvh{y}{t-i}}{\boldtheta}$ has time complexity 
$\mathcal{O}(h\, y\, \vartheta)$ because $W_i$ has size $h \times y$ and 
$\dvinline{\vvh{y}{t-i}}{\boldtheta}$ has size $y \times \vartheta$.
The summation of the obtained matrices has time complexity 
$\mathcal{O}(p\, h\, \vartheta)$.
\item[(iii)]
The multiplication between the matrix at obtained at step (i) and that obtained at step (ii) has time complexity $\mathcal{O}(h\, y\, \vartheta)$.
\end{itemize}
Hence, the leading-order term for executing line 2 is
$\mathcal{O}(p\, h\, y\, \vartheta)$.

\smallskip \noindent
The estimate of the time complexity is analogous for the equation at line 3. 
In this case, the time complexity term is
$\mathcal{O}(p\, h\, y\, \varphi)$, because for the leading order we are considering the same equation (cf.\ line 3) with $\boldphi$ instead of 
$\boldtheta$.
Hence, the total time complexity of line 2 and 3 is
$\mathcal{O}(p\, h\, y\, w)$.
As the algorithm needs to run for every element of the processed sequence of length $\tau$, the overall time complexity of RTRL is 
$\mathcal{O}(\tau\, p\, y\, h\, w)$.

\smallskip \noindent
Finally, the remaining operations are those at line 5. They can be neglected because: they are performed only once for every processed sequence; and they are matrix-vector (and not matrix-matrix) multiplications.

\smallskip
In terms of space complexity, at any time $t$ the algorithm only requires the storage of the previous $p$ Jacobian matrices
$\dvinline{\vvh{y}{t-i}}{\boldtheta}\vphantom{\Big[}$ and
$\dvinline{\vvh{y}{t-i}}{\boldphi}$, having size
$y \times \vartheta$ and $y \times \varphi$, respectively.
Thus, the space complexity of RTRL is
$\mathcal{O}(p\, y\, w)$.

\bigskip \noindent
(c) \underline{BPTT}.
To determine the time complexity of BPTT, we first need to deduce the time complexity associated with a single execution of the recursive function \texttt{backprop} in Algorithm 3 in Appendix \ref{sec:appendixALGO}.
We observe that:
\begin{itemize} \itemsep0em
	\item[(i)] At line 6, the computation of the vector 
	$\mathbf{g}_t\, V A^{(t)}\, W_k$ has time complexity 
	$\mathcal{O}(h\, y)$, analogously to step (i) of AAD.
	As this operation is performed $p$ times, we obtain 
	$\mathcal{O}(p\, h\, y)$.
	\item[(ii)] At line 8, the additions have complexity 
	$\mathcal{O}(w)$.
	As they are performed $p$ times, we obtain 
	$\mathcal{O}(p\, w)$.
	\item[(iii)] At line 11, the leading-order time complexities are
	$\mathcal{O}(h, \vartheta)$ and $\mathcal{O}(y, \varphi)$, respectively. The deduction is analogous to steps (ii) and (iii) of AAD.
\end{itemize}
We observe that step (i) has a time complexity of 
$\mathcal{O}(p,\ h\, y)$, where $p,\ h\, y < \vartheta$ by definition, cf.\ \eqref{eq:wtotalweights}.
As discussed in Section \ref{sec:model_spec}, $y$ and $p$ are assumed to be always less than $h$. 
Thus, on the one hand, the complexity 
$\mathcal{O}(h\, \vartheta + y\, \varphi)$ of step (iii) can be approximated from above as $\mathcal{O}(h\, w)$;
on the other, that of step (ii), namely
$\mathcal{O}(p\, w)$, is negligible compared to $\mathcal{O}(h\, w)$.
Hence, the leading-order time complexity for one execution of the recursive function \texttt{backprop} is $\mathcal{O}(h\, w)$.

\smallskip \noindent
To conclude the derivation of the time complexity, we need to compute how many times the \texttt{backprop} function is invoked.
This is equivalent to computing the number of the nodes in the unrolled tree representation presented in Section \ref{sec:thealgo}: this number is equal to $S_{\tau,p}$ (cf.\ also Appendix \ref{sec:appendixD}).

\noindent
This result follows from straightforward reasoning.
Every node with time-index $t$ is parent to 
one node with time-index $t-1$, 
one with time-index $t-2$,
$\dots$, and
one with time-index $t-p$.
Then, the number of nodes with time-index $j$ is equal to $f_{j,p}$, and the total number of nodes in the tree is equal to the partial sum 
$S_{\tau,p}$.
Thus, the total time complexity of BPTT is
$\mathcal{O}(S_{\tau, \, p}\, h\, w)$.

\smallskip
In terms of space complexity, BPTT is a backward-operating algorithm that, similarly to AAD, requires access to all the previous inputs $\vvv{x}{t}$, hidden states $\vvv{h}{t}$, and outputs $\vvh{y}{t}$.
For a sequence of length $\tau$, this yields a space complexity of 
$\mathcal{O}(\tau\, (x+h+y) )$
$\hfill\square$

\section{Learning Algorithms for RNN($p$)} \label{sec:appendixALGO}

In this appendix, we provide detailed formulations of the three learning algorithms for RNN($p$) models: AAD, RTRL, and BPTT.
These algorithms are straightforward and can be implemented in just a few lines of code in any programming language. However, they achieve significantly different time and space complexities, as discussed in Section \ref{ssec:tcsc}.

%
\begin{center}
\setlength{\algomargin}{0pt}
\begin{algorithm}
\caption{Adjoint Automatic Differentiation (AAD) for RNN($p$)}
\setstretch{1.20}
\vspace{0.1cm}
$\dv{L}{\vvh{y}{\tau}} = \pdv{L}{\vvh{y}{\tau}}$, \,
$\dv{\loss}{\boldtheta} = \mathbf{0}$, \,
$\dv{\loss}{\boldphi} = \mathbf{0}$
\;
\For{$t=\tau, \tau-1, \dots, 1$}{
	$
	\dv{L}{\vvh{y}{t}} =
	\sum_{k=1}^{\min(p, \tau-t)}
		\dv{L}{\vvh{y}{t+k}} \,
		V A^{(t+k)}\, W_k$
	\;
	$
	\dv{\loss}{\boldtheta}
	\, \, +\!\!= \,\,
		\dv{\loss}{ \vvh{y}{t} }
		~
		V A^{(t)} \,
		\pdv{ \vvv{a}{t} }{\boldtheta}
	$
	\;
	$
	\dv{\loss}{\boldphi}
	\, \, +\!\!= \,\,
		\dv{\loss}{ \vvh{y}{t} }
		~
		\pdv{ \vvh{y}{t} }{\boldphi}
	$
	}
\end{algorithm}

\vspace{-0.25cm}
\setlength{\algomargin}{0pt}
\begin{algorithm}
	\setstretch{1.20}
	\caption{Real-Time Recurrent Learning (RTRL) for RNN($p$)}
	\For{$t=1, 2, \dots, \tau$}{
	$
	\dv{\vvh{y}{t}}{\boldtheta}
	=
	V
	A^{(t)}
	\left(
		\pdv{\vvv{a}{t}}{\boldtheta}
		+
		\sum_{k=1}^{\min(p, t-1)}
		W_k
		\dv{\vvh{y}{t-k}}{\boldtheta}
	\right)
	$
	\\[0.1cm]
	$
	\dv{\vvh{y}{t}}{\boldphi}
	=
	\,
	\pdv{\vvh{y}{t}}{\boldphi}
	\,+\,
	V
	A^{(t)} \,
	\sum_{k=1}^{\min(p, t-1)}
	 	W_k
		\dv{\vvh{y}{t-k}}{\boldphi}$
	}
	$\dv{\loss}{\boldtheta} =
	\pdv{\loss}{ \vvh{y}{\tau} }
	~
	\dv{ \vvh{y}{\tau} }{\boldtheta}
	, ~
	\dv{\loss}{\boldphi}
	=
	\pdv{\loss}{ \vvh{y}{\tau} }
	\dv{ \vvh{y}{\tau} }{\boldphi}$ \\[0.2cm]
	%
\end{algorithm}

\vspace{-0.25cm}
\setlength{\algomargin}{0pt}
\begin{algorithm}
	\setstretch{1.20}
	\caption{Backpropagation Through Time (BPTT) for RNN($p$)}
	\Function{backprop($\mathbf{g}_t, t$)}{
		\CommentSty{// $\mathbf{g}_t$:\ partial derivative of $L$ w.r.t.\ $\vvh{y}{t}$ on the current path}\\
		\CommentSty{// $t$:\ current time step (counting backward from final step $\tau$)} \\[0.1cm]
		$\nabla_\theta = \mathbf{0}$,\quad
		$\nabla_\varphi = \mathbf{0}$ \\[0.0cm]
			\For{$k = 1, 2, \dots, \min(p, t - 1)~$}{
				$\mathbf{g}_{t-k} = \mathbf{g}_t \, V \, A^{(t)} \, W_k$ \\[0.1cm]
				$[\mathbf{d}_1, \mathbf{d}_2] =$
				backprop($\mathbf{g}_{t-k},\; t - k$) \\[0.1cm]
				$\nabla_\theta \mathrel{+}= \mathbf{d}_1$,\quad
				$\nabla_\varphi \mathrel{+}= \mathbf{d}_2$
			}
			$\nabla_\theta
			\, +\!\!=  \,
			\mathbf{g}_t\, V A^{(t)} \,\pdv{\vvv{a}{t}}{\boldtheta}$, \quad
			$\nabla_\varphi
			\, +\!\!=  \,
			\mathbf{g}_t\, \pdv{\vvh{y}{t}}{\boldphi}$  \\[0.15cm]
		\Return $\nabla_\theta,\; \nabla_\varphi$
	}
	\vspace{0.2cm}
	\CommentSty{// Main call:~start from final time step $\tau$} \\
	$\left[ 
		\mygradient{\loss}{\boldtheta}, 
		\mygradient{\loss}{\boldphi}
	\right] =$ backprop$\left(\pdv{L}{\vvh{y}{\tau}},\; \tau\right)$
\end{algorithm}
\end{center}

\clearpage
\newpage

In addition, in Figure \ref{fig:chain} we provide two graphical representations of the gradient backpropagation through the unrolled network, corresponding to the RNN($1$) case \textit{(above)} and the RNN($2$) case \textit{(below)}.
These plots are analogous to those in Figure \ref{fig:figure_bptt}, but they highlight the contributions to the gradients: in blue for $\dvinline{\loss}{\boldtheta}$ and in red for $\dvinline{\loss}{\boldphi}$. Notably, in the RNN($2$) case, we observe the increasing number of contributions as the network is unrolled.\\

\begin{figure*}[!h]
	\centering
	\resizebox{0.95\linewidth}{!}{
\begin{tikzpicture}[scale=.9, ->, thick, draw=black!75, shorten <= 5pt, shorten >= 5pt]
    
    \draw (16,5.10) circle(0.0) node[] (L) {\small{$\loss$}};
    \draw[rounded corners] (15.15, 4.75) rectangle ++(1.7, 0.7) {};
    
    \draw (0,3) circle(0.0) node[] (Y-1) {$\vvh{y}{1}$};
    \draw[rounded corners] (-0.85, 2.65) rectangle ++(1.7, 0.7) {};
    \draw (0,1) circle(0.0) node[] (A-1) {$\vvv{a}{1}$};
    \draw[rounded corners] (-0.85, 0.65) rectangle ++(1.7, 0.7) {};    
       
    \draw (4,3) circle(0.0) node[] (Y-2) {$\vvh{y}{2}$};
    \draw[rounded corners] (3.15, 2.65) rectangle ++(1.7, 0.7) {};
    \draw (4,1) circle(0.0) node[] (A-2) {$\vvv{a}{2}$};
    \draw[rounded corners] (3.15, 0.65) rectangle ++(1.7, 0.7) {};         
  
    \draw (8,3) circle(0.0) node[] (Y-3) {$\vvh{y}{\mathbf{\cdot}}$};
    \draw[rounded corners] (7.15, 2.65) rectangle ++(1.7, 0.7) {};
    \draw (8,1) circle(0.0) node[] (A-3) {$\vvv{a}{\mathbf{\cdot}}$};
    \draw[rounded corners] (7.15, 0.65) rectangle ++(1.7, 0.7) {};   
            
    \draw (12,3) circle(0.0) node[] (Y-4) {$\vvh{y}{\tau-1}$};
    \draw[rounded corners] (11.15, 2.65) rectangle ++(1.7, 0.7) {};
    \draw (12,1) circle(0.0) node[] (A-4) {$\vvv{a}{\tau-1}$};
    \draw[rounded corners] (11.15, 0.65) rectangle ++(1.7, 0.7) {};   
    
    \draw (16,3) circle(0.0) node[] (Y-5) {$\vvh{y}{\tau}$};
    \draw[rounded corners] (15.15, 2.65) rectangle ++(1.7, 0.7) {};
    \draw (16,1) circle(0.0) node[] (A-5) {$\vvv{a}{\tau}$};
    \draw[rounded corners] (15.15, 0.65) rectangle ++(1.7, 0.7) {};

	\draw[->, CadetBlue, dashed] (0, 2.7) -- (0, 1.3);
    \draw[CadetBlue, dashed] (0,1.9) circle(0.0) node[right] {$V A^{(1)}$};

	\draw[->, CadetBlue, dashed] (4, 2.7) -- (4, 1.3);
    \draw[CadetBlue, dashed] (4,1.9) circle(0.0) node[right] {$V A^{(2)}$};

	\draw[->, CadetBlue, dashed] (8, 2.7) -- (8, 1.3);
    \draw[CadetBlue, dashed] (8,1.9) circle(0.0) node[right] {$V A^{(\mathbf{\cdot})}$};

	\draw[->, CadetBlue, dashed] (11.9, 2.7) -- (11.9, 1.3);
    \draw[CadetBlue, dashed] (11.9,1.9) circle(0.0) node[right] {$V A^{(\tau-1)}$};

	\draw[->, CadetBlue, dashed] (16, 2.7) -- (16, 1.3);
    \draw[CadetBlue, dashed] (16,1.9) circle(0.0) node[right] {$V A^{(\tau)}$};

	\draw[->, CadetBlue, dashed] (16, 4.8) -- (16, 3.3);
    \draw[CadetBlue, dashed] (16,4) circle(0.0) node[right] {$\pdv{\loss}{\vvh{y}{\tau}}$};

	\draw[->, CadetBlue, dashed] (15.20, 1.55) -- (12.9, 3.05);
	\draw[->, CadetBlue, dashed] (11.20, 1.55) -- ( 8.9, 3.05);
	\draw[->, CadetBlue, dashed] ( 7.20, 1.55) -- ( 4.9, 3.05);
	\draw[->, CadetBlue, dashed] ( 3.20, 1.55) -- ( 0.9, 3.05);


    \draw[CadetBlue, dashed] ( 1.9,2.6) circle(0.0) node[right] {$W_1$};
    \draw[CadetBlue, dashed] ( 5.9,2.6) circle(0.0) node[right] {$W_1$};
    \draw[CadetBlue, dashed] ( 9.9,2.6) circle(0.0) node[right] {$W_1$};
    \draw[CadetBlue, dashed] (13.9,2.6) circle(0.0) node[right] {$W_1$};



    
    \draw[red, double, dashed] (Y-1) -- (-1.35,4) node 
        [above left = -0.25cm] {$\pdv{\vvh{y}{1}}{\boldphi}$};
    \draw[red, double, dashed] (Y-2) -- ( 2.65,4) node 
        [above left = -0.25cm] {$\pdv{\vvh{y}{2}}{\boldphi}$};
    \draw[red, double, dashed] (Y-3) -- ( 6.65,4) node 
        [above left = -0.25cm] {$\pdv{\vvh{y}{\mathbf{\cdot}}}{\boldphi}$};
    \draw[red, double, dashed] (Y-4) -- (10.65,4) node 
        [above left = -0.25cm] {$\pdv{\vvh{y}{\tau-1}}{\boldphi}$};
    \draw[red, double, dashed] (Y-5) -- (14.65,4) node 
        [above left = -0.25cm] {$\pdv{\vvh{y}{\tau}}{\boldphi}$};

    \draw[blue, double] (A-1) -- (-1.35, 0.1) node 
        [below left = -0.25cm] {$\pdv{\vvv{a}{1}}{\boldtheta}$};
    \draw[blue, double] (A-2) -- ( 2.65, 0.1) node 
        [below left = -0.25cm] {$\pdv{\vvv{a}{2}}{\boldtheta}$};
    \draw[blue, double] (A-3) -- ( 6.65, 0.1) node 
        [below left = -0.25cm] {$\pdv{\vvv{a}{\mathbf{\cdot}}}{\boldtheta}$};
    \draw[blue, double] (A-4) -- (10.65, 0.1) node 
        [below left = -0.25cm] {$\pdv{\vvv{a}{\tau-1}}{\boldtheta}$};
    \draw[blue, double] (A-5) -- (14.65, 0.1) node
            [below left = -0.25cm] {$\pdv{\vvv{a}{\tau}}{\boldtheta}$};

\begin{scope}[yshift=-22.0cm]
    
    
    \draw (8,19.25) circle(0.0) node[] (L) {\small{$\loss$}};
    \draw[rounded corners] (7.35, 18.90) rectangle ++(1.3, 0.7) {};

    
    \draw (8,17) circle(0.0) node[] (AY-1) {$\vvh{y}{\tau}$};
    \draw[rounded corners] (7.35, 16.65) rectangle ++(1.3, 0.7) {};
    \draw (8,15) circle(0.0) node[] (A-1) {$\vvv{a}{\tau}$};
    \draw[rounded corners] (7.35, 14.65) rectangle ++(1.3, 0.7) {};    
    
    \draw [] (3,14)  circle (0.0) node[] (BY-1) {$\vvh{y}{\tau-1}$};
    \draw[rounded corners] (2.25, 13.65) rectangle ++(1.5, 0.7) {};
    \draw [] (3,12)  circle (0.0) node[] (B-1) {$\vvv{a}{\tau-1}$};
    \draw[rounded corners] (2.25, 11.65) rectangle ++(1.5, 0.7) {};
    
    \draw [] (13,14) circle (0.0) node[] (BY-2) {$\vvh{y}{\tau-2}$};
    \draw[rounded corners] (12.25, 13.65) rectangle ++(1.5, 0.7) {};
    \draw [] (13,12) circle (0.0) node[] (B-2) {$\vvv{a}{\tau-2}$};
    \draw[rounded corners] (12.25, 11.65) rectangle ++(1.5, 0.7) {};

    \draw [] (1.00,10)  circle (0.0) node[] (CY-1) {$\vvh{y}{\tau-2}$};
    \draw[rounded corners] (0.25, 9.65) rectangle ++(1.5, 0.7) {};
    \draw [] (5.00,10)  circle (0.0) node[] (CY-2) {$\vvh{y}{\tau-3}$};
    \draw[rounded corners] (4.25, 9.65) rectangle ++(1.5, 0.7) {};
    \draw [] (11.00,10) circle (0.0) node[] (CY-3) {$\vvh{y}{\tau-3}$};
    \draw[rounded corners] (10.25, 9.65) rectangle ++(1.5, 0.7) {};
    \draw [] (15.00,10) circle (0.0) node[] (CY-4) {$\vvh{y}{\tau-4}$};
    \draw[rounded corners] (14.25, 9.65) rectangle ++(1.5, 0.7) {};

    \draw [] (1.00,8)  circle (0.0) node[] (C-1) {$\vvv{a}{\tau-2}$};
    \draw[rounded corners] (0.25, 7.65) rectangle ++(1.5, 0.7) {};
    \draw [] (5.00,8)  circle (0.0) node[] (C-2) {$\vvv{a}{\tau-3}$};
    \draw[rounded corners] (4.25, 7.65) rectangle ++(1.5, 0.7) {};
    \draw [] (11.00,8) circle (0.0) node[] (C-3) {$\vvv{a}{\tau-3}$};
    \draw[rounded corners] (10.25, 7.65) rectangle ++(1.5, 0.7) {};
    \draw [] (15.00,8) circle (0.0) node[] (C-4) {$\vvv{a}{\tau-4}$};
    \draw[rounded corners] (14.25, 7.65) rectangle ++(1.5, 0.7) {};

    \draw [] (0, 5.5)  circle (0.0) node[] (D-1) {$\vvh{y}{\tau-3}$};
    \draw[rounded corners] (-0.75, 5.15) rectangle ++(1.5, 0.7) {};
    \draw [] (2, 5.5)  circle (0.0) node[] (D-2) {$\vvh{y}{\tau-4}$};
    \draw[rounded corners] (1.25, 5.15) rectangle ++(1.5, 0.7) {};
    \draw [] (4, 5.5)  circle (0.0) node[] (D-3) {$\vvh{y}{\tau-4}$};
    \draw[rounded corners] (3.25, 5.15) rectangle ++(1.5, 0.7) {};
    \draw [] (6, 5.5)  circle (0.0) node[] (D-4) {$\vvh{y}{\tau-5}$};
    \draw[rounded corners] (5.25, 5.15) rectangle ++(1.5, 0.7) {};
    \draw [] (10, 5.5)  circle (0.0) node[] (D-5) {$\vvh{y}{\tau-4}$};
    \draw[rounded corners] (9.25, 5.15) rectangle ++(1.5, 0.7) {};
    \draw [] (12, 5.5) circle (0.0) node[] (D-6) {$\vvh{y}{\tau-5}$};
    \draw[rounded corners] (11.25, 5.15) rectangle ++(1.5, 0.7) {};
    \draw [] (14, 5.5) circle (0.0) node[] (D-7) {$\vvh{y}{\tau-5}$};
    \draw[rounded corners] (13.25, 5.15) rectangle ++(1.5, 0.7) {};
    \draw [] (16, 5.5) circle (0.0) node[] (D-8) {$\vvh{y}{\tau-6}$};
    \draw[rounded corners] (15.25, 5.15) rectangle ++(1.5, 0.7) {};

    \path[CadetBlue, dashed] (L) edge[] node 
        [right] {$\pdv{\loss}{\vvh{y}{\tau}}$} (AY-1);
    \path[CadetBlue, dashed] (AY-1) edge[] node 
        [right] {$V A^{(\tau)}$} (A-1);

    \path[CadetBlue, dashed] (A-1) edge[] node [above left] {$W_{1}$} (BY-1);
    \path[CadetBlue, dashed] (A-1) edge[] node [above right] {$W_{2}$} (BY-2);

    \path[CadetBlue, dashed] (BY-1) edge[] node 
        [right] {$V A^{(\tau-1)}$} (B-1);
    \path[CadetBlue, dashed] (BY-2) edge[] node 
        [left] {$V A^{(\tau-2)}$} (B-2);

    \path[CadetBlue, dashed] (B-1) edge[] node 
    	[above left= 0.1cm,anchor=320] {$W_{1}$} (CY-1);
    \path[CadetBlue, dashed] (B-1) edge[] node 
    	[above right= 0.1cm, anchor=220] {$W_{2}$} (CY-2);
    \path[CadetBlue, dashed] (B-2) edge[] node 
    	[above left= 0.1cm, anchor=320] {$W_{1}$} (CY-3);
    \path[CadetBlue, dashed] (B-2) edge[] node 
    	[above right= 0.1cm,anchor=220] {$W_{2}$} (CY-4);

    \path[CadetBlue, dashed] (CY-1) edge[] node 
        [right] {$V A^{(\tau-2)}$} (C-1);
    \path[CadetBlue, dashed] (CY-2) edge[] node 
        [left] {$V A^{(\tau-3)}$} (C-2);
    \path[CadetBlue, dashed] (CY-3) edge[] node 
        [right] {$V A^{(\tau-3)}$} (C-3);
    \path[CadetBlue, dashed] (CY-4) edge[] node 
        [left] {$V A^{(\tau-4)}$} (C-4);
       
    \path[CadetBlue, dashed] (C-1) edge[] node 
        [above left= 0.1cm,anchor=340] {$W_{1}$} (D-1);
    \path[CadetBlue, dashed] (C-1) edge[] node 
        [above right= 0.1cm,anchor=200] {$W_{2}$} (D-2);
    \path[CadetBlue, dashed] (C-2) edge[] node 
        [above left= 0.1cm,anchor=340] {$W_{1}$} (D-3);
    \path[CadetBlue, dashed] (C-2) edge[] node 
        [above right= 0.1cm,anchor=200] {$W_{2}$} (D-4);
    \path[CadetBlue, dashed] (C-3) edge[] node 
        [above left= 0.1cm,anchor=320] {$W_{1}$} (D-5);
    \path[CadetBlue, dashed] (C-3) edge[] node 
        [above right= 0.1cm,anchor=220] {$W_{2}$} (D-6);
    \path[CadetBlue, dashed] (C-4) edge[] node 
        [above left= 0.1cm,anchor=320] {$W_{1}$} (D-7);
    \path[CadetBlue, dashed] (C-4) edge[] node 
        [above right= 0.1cm,anchor=220] {$W_{2}$} (D-8);

    
    \draw[red, double, dashed] (AY-1) -- (6.75, 18) node 
        [above left = -0.25cm] {$\pdv{\vvh{y}{\tau}}{\boldphi}$};
    \draw[blue, double] (A-1) -- (6.75, 16) node 
        [above left = -0.25cm] {$\pdv{\vvv{a}{\tau}}{\boldtheta}$};

    \draw[red, double, dashed] (BY-1) -- (1.75, 15) node 
        [above left = -0.25cm] {$\pdv{\vvh{y}{\tau-1}}{\boldphi}$};
    \draw[blue, double] (B-1) -- (1.75, 13) node 
        [above left = -0.25cm] {$\pdv{\vvv{a}{\tau-1}}{\boldtheta}$};
    \draw[red, double, dashed] (BY-2) -- (14.25, 15) node 
        [above right = -0.25cm] {$\pdv{\vvh{y}{\tau-2}}{\boldphi}$};
    \draw[blue, double] (B-2) -- (14.25, 13) node 
        [above right = -0.25cm] {$\pdv{\vvv{a}{\tau-2}}{\boldtheta}$};
 
    \draw[red, double, dashed] (CY-1) -- (-0.5,11) node 
        [above left = -0.45cm] {$\pdv{\vvh{y}{\tau-2}}{\boldphi}$};
    \draw[red, double, dashed] (CY-2) -- (6.5,11) node 
        [above right = -0.45cm] {$\pdv{\vvh{y}{\tau-3}}{\boldphi}$};
    \draw[red, double, dashed] (CY-3) -- (9.5,11) node 
        [above left = -0.45cm] {$\pdv{\vvh{y}{\tau-3}}{\boldphi}$};
    \draw[red, double, dashed] (CY-4) -- (16.5,11) node 
       [above right = -0.45cm] {$\pdv{\vvh{y}{\tau-4}}{\boldphi}$};

    \draw[blue, double] (C-1) -- (-0.5,9) node 
        [above left  = -0.45cm] {$\pdv{\vvv{a}{\tau-2}}{\boldtheta}$};
    \draw[blue, double] (C-2) -- (6.5,9) node 
        [above right = -0.45cm] {$\pdv{\vvv{a}{\tau-3}}{\boldtheta}$};
    \draw[blue, double] (C-3) -- (9.5,  9) node 
        [above left  = -0.45cm] {$\pdv{\vvv{a}{\tau-3}}{\boldtheta}$};
    \draw[blue, double] (C-4) -- (16.5, 9) node 
        [above right = -0.45cm] {$\pdv{\vvv{a}{\tau-4}}{\boldtheta}$};


	\draw[blue, double] (11.4,19.25) -- (12.4,19.25) node 
		[right=0.0cm] {\small{Contribution to $\boldtheta$-gradient}};
	\draw[red, double, dashed] (11.4,18.75) -- (12.4,18.75) node 
		[right=0.0cm] {\small{Contribution to $\boldphi$-gradient}};
	\draw[CadetBlue, dashed] (11.4,18.25) -- (12.4,18.25) node
		[right=0.0cm] {\small{Backpropagation flow}};
    \draw[rounded corners, black] (11.33, 17.90) rectangle ++(6.5, 1.75) {};	 

\end{scope}

    \node[] at (-1.0,   5.60) {\textbf{RNN($1$)}};
    \node[] at (-1.0,  -2.50) {\textbf{RNN($2$)}};

\end{tikzpicture}
	}
	\smallskip
	\caption{\small{
	BPTT for an RNN($1$) \textit{(above)} and for an RNN($2$) \textit{(below)}.
	Gray dashed arrows represent the backpropagation flow,
	blue double solid arrows represent the contributions to the
	gradient $\dvinline{\loss}{\boldtheta}$, and
	red double dashed arrows represent the contributions to the
	gradient $\dvinline{\loss}{\boldphi}$.
	}}
	\label{fig:chain}
\end{figure*}

\clearpage
\newpage

\section{Additional Analyses and Robustness Checks} \label{sec:appendixC}

This appendix presents extended experimental results that complement the main findings of Section~\ref{sec:energy}. We report performance on additional test years, evaluate forecast ensemble methods, and provide further robustness checks for the RNN($p$) architecture. For brevity, the analyses focus on the New England dataset; analogous results are found for the London dataset and are available upon request. At the end of this appendix, we also provide a detailed specification of the benchmark models used in the study.

\subsection{Additional held-out test sets}
\label{subsec:appendixC:extended_years}

We evaluate probabilistic forecasting performance on two additional held-out test years (2013 and 2014). The results confirm that the RNN($\{1,2,24\}$) architecture consistently outperforms all benchmarks across both years, demonstrating stable performance beyond the 2012 test set analysed in Section~\ref{sec:energy}.

\begin{table}[!h]
\centering
\resizebox{\textwidth}{!}{%
\begin{tabular}{ l | c c c c || c c c c }
\toprule
& \multicolumn{4}{c||}{\textbf{2013}} & \multicolumn{4}{c}{\textbf{2014}} \\
\cmidrule(lr){2-5} \cmidrule(lr){6-9}
Model & MAPE [\%] & RMSE [MWh] & APL [MWh] & NLL & MAPE [\%] & RMSE [MWh] & APL [MWh] & NLL \\
\midrule
Naïve & 6.14 & 1334.91 & 330.01 & 8.37 & 
	    6.25 & 1210.05 & 317.81 & 8.38 \\
ARX & 5.03 & 1022.08 & 275.73 & 8.28 & 
	  5.45 & 1023.34 & 280.44 & 8.32 \\
SARIMAX & 5.01 & 985.17 & 270.68 & 8.27 & 
		  5.32 & 964.96 & 267.90 & 8.26 \\
LightGBM & 3.11 $\pm$ 0.02 & 632.35 $\pm$ 2.99 & 171.92 $\pm$ 1.03 & 7.82 $\pm$ 0.01 & 
			2.95 $\pm$ 0.01 & 569.95 $\pm$ 1.91 & 156.41 $\pm$ 0.59 & 7.73 $\pm$ 0.01 \\
FNN & 2.71 $\pm$ 0.04 & 584.23 $\pm$ 14.68 & 148.61 $\pm$ 2.26 & 7.60 $\pm$ 0.01 & 
		3.03 $\pm$ 0.06 & 567.31 $\pm$ 10.62 & 154.63 $\pm$ 2.97 & 7.73 $\pm$ 0.02 \\
RNN($1$) & 2.30 $\pm$ 0.01 & 456.40 $\pm$ 2.32 & 123.36 $\pm$ 0.61 & 7.46 $\pm$ 0.01 & 2.46 $\pm$ 0.03 & 459.92 $\pm$ 4.18 & 126.40 $\pm$ 1.28 & 7.55 $\pm$ 0.01 \\
RNN($\{1,2,24\}$) & $\mathbf{2.14 \pm 0.01}$ & $\mathbf{423.06 \pm 2.60}$ & $\mathbf{115.14 \pm 0.60}$ & $\mathbf{7.40 \pm 0.01}$ & $\mathbf{2.29 \pm 0.02}$ & $\mathbf{433.34 \pm 2.24}$ & $\mathbf{118.62 \pm 0.77}$ & $\mathbf{7.49 \pm 0.01}$ \\
LSTM & 2.18 $\pm$ 0.05 & 440.80 $\pm$ 9.64 & 119.40 $\pm$ 2.72 & 7.43 $\pm$ 0.03 & 
	   2.50 $\pm$ 0.05 & 467.04 $\pm$ 8.51 & 129.22 $\pm$ 2.26 & 7.59 $\pm$ 0.03 \\
Transformer & 
	2.20 $\pm$ 0.03 & 447.74 $\pm$ 7.31 & 119.97 $\pm$ 1.85 & 7.43 $\pm$ 0.02 & 
	2.41 $\pm$ 0.04 & 459.52 $\pm$ 8.71 & 124.92 $\pm$ 2.32 & 7.53 $\pm$ 0.03 \\
\midrule
\end{tabular}
}
\caption{Comparison of RNN($p$) with benchmarks on the New England dataset: probabilistic forecasting results on the additional test sets (2013 and 2014). RNN($\{1,2,24\}$) maintains exceptional accuracy across both periods.}
\label{tab:2014_results}
\end{table}

\subsection{Ensemble Performance}
\label{subsec:appendixC:ensembles}

We investigate \textit{forecast ensembling} by averaging, for every hour, the predictions produced by the RNN($p$) models across the 10 independent runs. As shown in Table~\ref{tab:ensembles_results}, the inclusion of the full multi-lag structure $\{1,2,24\}$ yields substantial improvements over the vanilla RNN($1$) ensemble.

\begin{table}[!h]
\centering
\begin{tabular}{ c | c c c c}
\midrule
Ensemble Configuration & MAPE [\%] & RMSE [MWh] & APL [MWh] & NLL \\
\midrule
Ensemble($\{1\}$) & 2.20 & 463.70 & 119.48 & 7.45 \\
Ensemble($\{1,2\}$) & 2.14 & 454.05 & 116.55 & 7.42 \\
Ensemble($\{1,2,24\}$) & \textbf{2.02} & \textbf{427.14} & \textbf{108.99} & \textbf{7.36} \\
\midrule
\end{tabular}
\caption{
Performance of ensemble models combining the 10 RNN($p$) instances for probabilistic forecasting on the New England dataset (2012 test set). The best results are highlighted in bold.
}
\label{tab:ensembles_results}
\end{table}

\subsection{Effect of Seasonality Removal}

In Table~\ref{tab:no_deseasonalization}, we examine the importance of the deseasonalisation preprocessing step by training RNN($p$) models on raw, non-deseasonalised data. Performance degrades substantially across all lag configurations, confirming that explicit seasonal handling is crucial for optimal results.

\begin{table}[H]
\centering
\begin{tabular}{ c | c c c c }
\midrule
Lags $p$ & MAPE [\%] & RMSE [MWh] & APL [MWh] & NLL \\
\midrule
$\{1\}$        & 3.12 $\pm$ 0.03 & 535.20 $\pm$ 9.10 & 138.50 $\pm$ 2.80 & 7.82 $\pm$ 0.03 \\
$\{1,2\}$      & 2.97 $\pm$ 0.02 & 510.75 $\pm$ 8.50 & 132.10 $\pm$ 2.60 & 7.70 $\pm$ 0.02 \\
$\{1,2,24\}$   & 2.83 $\pm$ 0.02 & 485.90 $\pm$ 7.90 & 126.30 $\pm$ 2.40 & 7.59 $\pm$ 0.02 \\
\midrule
\end{tabular}
\caption{
Performance of RNN($p$) models \textit{without} seasonal preprocessing on the 2012 test set. All metrics show significant deterioration compared to the corresponding deseasonalised results in the main paper (cf.\ Table~\ref{tab:prob_results}).
}
\label{tab:no_deseasonalization}
\end{table}

\subsection{Sensitivity to Lag Specification}

\begin{figure}[H]
	\centering
	\includegraphics[width=0.75\linewidth, trim={0.2cm 2.5cm 0.2cm 1.2cm}, clip]{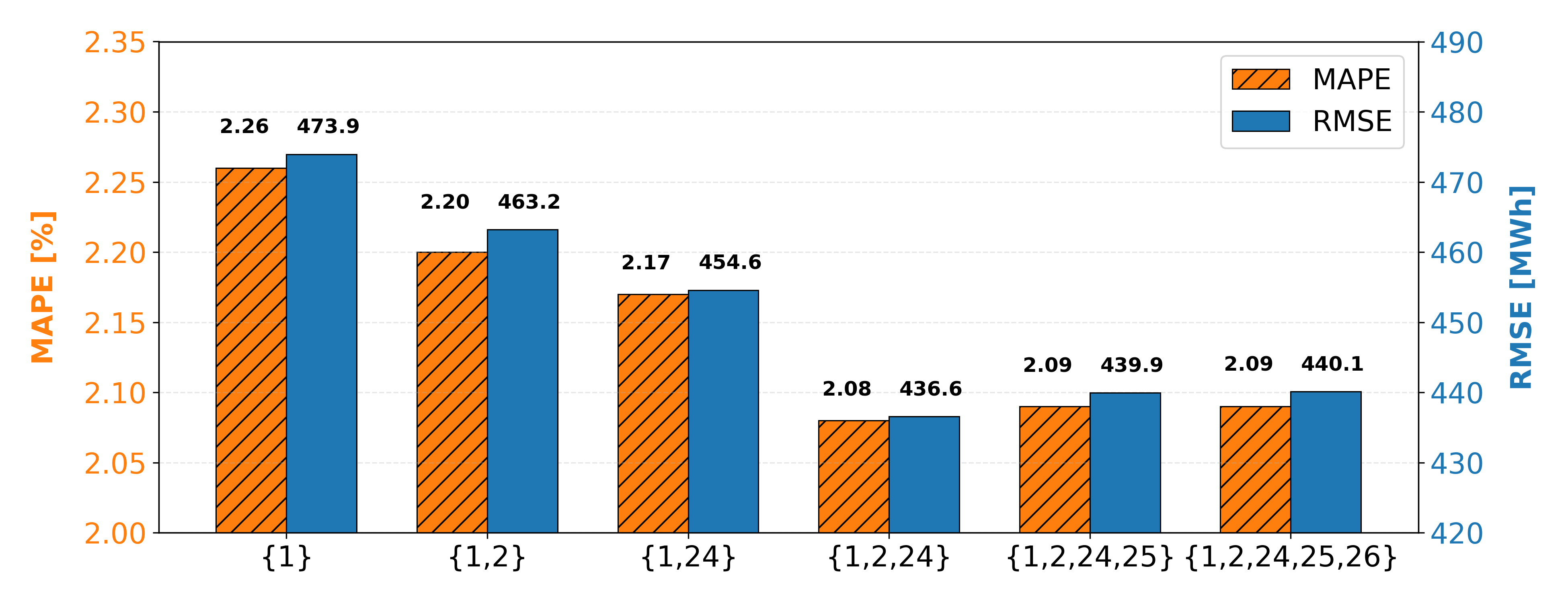}
	\caption{Sensitivity of RNN($p$) performance to lag specification for probabilistic forecasting on the New England 2012 test set. The configuration $\{1,2,24\}$ provides an effective trade-off between complexity and accuracy.}
\label{fig:lag_sensitivity}
\end{figure}

Figure~\ref{fig:lag_sensitivity} explores the impact of different lag specifications beyond the three primary configurations. Performance generally improves with the inclusion of both short-term ($p=1,2$) and daily ($p=24$) lags. The results remain robust even when additional feedbacks are incorporated.

\subsection{Validity of Gaussian Likelihood Assumption}

\begin{figure}[H]
	\centering
	\includegraphics[width=0.43\linewidth, trim={0.2cm 0.2cm 0.2cm 1.0cm}, clip]{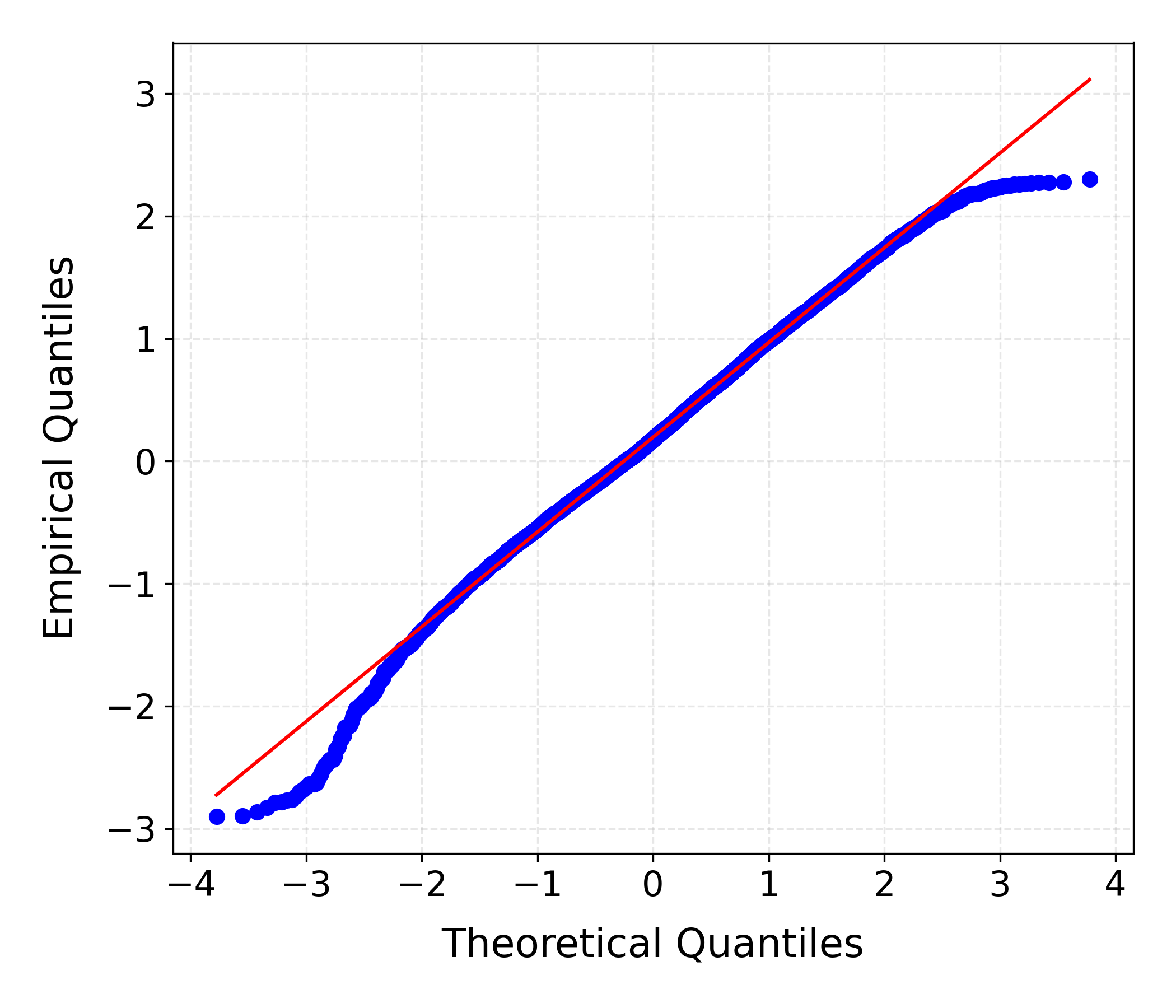}
	\caption{QQ-plot of standardised residuals for the RNN($\{1,2,24\}$) model's probabilistic forecasts on the New England 2012 test set. The close alignment with the diagonal suggests the Gaussian likelihood is appropriate.}
\label{fig:qq_gaussianity}
\end{figure}

A QQ-plot of standardised residuals (Figure~\ref{fig:qq_gaussianity}) confirms that the Gaussian distribution assumption for the probabilistic forecasts is reasonable, with minor deviations primarily in the tails.

\subsection{Robustness to Loss Function}

Finally, in the point forecasting case, we retrain models using Mean Absolute Error (MAE) as the loss function to verify that results are not specific to the MSE loss. Table~\ref{tab:mae_loss} shows that relative performance rankings remain unchanged, confirming the architectural advantage of RNN($\{1,2,24\}$).

\begin{table}[H]
\centering
\begin{tabular}{ c | c c }
	\midrule
	Lags $p$ & MAPE [\%] & RMSE [MWh] \\
	\midrule
	$\{1\}$ & 2.15 $\pm$ 0.01 & 454.98 $\pm$ 2.64 \\
	$\{1,2\}$ & 2.10 $\pm$ 0.02 & 442.40 $\pm$ 3.13 \\
	$\{1,2,24\}$ & \textbf{2.06 $\pm$ 0.01} & \textbf{430.44 $\pm$ 2.32} \\
	\midrule
\end{tabular}
\caption{
Point forecasting performance using MAE as the training loss on the New England dataset (2012 test set). The ranking of models is preserved, indicating robustness to the choice of loss function.
}
\label{tab:mae_loss}
\end{table}

\quad\smallskip
\subsection*{Benchmark Model Descriptions}

\medskip
{\small
\renewcommand{\arraystretch}{1.2} 
\begin{tabular}{@{}p{0.15\textwidth}p{0.8\textwidth}@{}}
\textbf{Naïve:} & For each hour, the point forecast is computed as the average load recorded in the same month, on the same weekday, and at the same hour over the in-sample period. The predictive uncertainty is derived from the standard deviation of these observations. \\
\textbf{ARX:} & ARX model that uses the same autoregressive lags as the reference RNN($p$), i.e.\ $\{1, 2, 24\}$.
Implemented in Python using \texttt{statsmodels}. \\
\textbf{SARIMAX:} & Seasonal ARIMA model with order $(p,d,q)=(1,0,1)$ and seasonal order $(P,D,Q,s)=(1,0,0,24)$. Implemented in Python using \texttt{statsmodels}. \\
\textbf{LightGBM:} & LightGBM with 1000 trees (50 leaves, learning rate 0.01) and 80\% feature/data sub‑sampling. Implemented in Python using \texttt{lightgbm}. \\
\textbf{FNN:} & Sequential \texttt{TensorFlow} model with two Dense layers. It features ReLU activation, 15 hidden neurons, learning rate 0.001 and batch size 64 (New England); ReLU activation, 10 hidden neurons,
learning rate 0.001, batch size 64 (London). \\
\textbf{LSTM:} & Sequential \texttt{TensorFlow} model with one LSTM layer followed by a Dense layer. It features sigmoid activation, 15 units, learning rate 0.005 and batch size 64 (New England); sigmoid activation, 10 units, learning rate 0.005, batch size 32 (London). \\
\textbf{Transformer:} & SingleDeT implemented using \texttt{TensorFlow}, featuring a 2‑layer encoder‑decoder stack with 32‑dimensional embeddings and 4 attention heads. Trained with a learning rate of 0.001 and a batch size of 64, using a 48‑hour input window. \\
\end{tabular}
\renewcommand{\arraystretch}{1.0}  
}

\clearpage
\newpage

\section{The $p$-bonacci sequences} \label{sec:appendixD}

A $p$-bonacci sequence is a generalisation of the classical Fibonacci sequence, where each term is defined as the sum of the previous $p$ terms, rather than just the previous two \citep[see, e.g.,][]{miles1960generalized}. 
For any natural integer $p$, the $p$-bonacci sequence $( f_{j,\,p} )_{j}$ is recursively defined by the following relationship:
\begin{equation}\label{eq:p-bonacci}
	\begin{cases}
		\displaystyle
		f_{j,\,p} \, := \hspace{-0.3cm}
		\sum_{k = 1}^{ \min(p, j-1) } \hspace{-0.1cm} f_{j-k, \, p}
		\\[0.5cm]
		\displaystyle
		f_{1,\,p} := 1
		\vphantom{\sum_{}^{n-1}}
	\end{cases}
\end{equation}
analogous to (1) in \citet[][]{miles1960generalized}.
Notably, when \( p = 1 \), the sequence is constant: \( f_{t,\,1} = 1 \) for all \( t \geq 1 \). For \( p = 2 \), the Fibonacci sequence is recovered, while for \( p = 3 \), the resulting sequence is known as the Tribonacci sequence. These sequences serve as valuable tools in various algorithmic contexts, particularly in the study of recursive structures and additive combinatorics \citep[see, e.g.,][]{cormen2022introduction}.


The $j$-th partial sum of a $p$-bonacci sequence, denoted as $S_{j,\,p}$, is defined as
\begin{equation}\label{eq:sump-bonacci}
	S_{j,p} \, := \,
	\sum_{k = 1}^{j} \, f_{k,\,p} \,,
\end{equation}
An important property of the partial sums of $p$-bonacci sequences (with $p\geq2$) is that they grow exponentially with respect to $j$, as we prove in the following lemma.

\begin{lemma} \label{lemma:p-bonacci}
	Let $S_{j,p}$ be the $j{\textnormal{-th}}$ partial sum of a $p$-bonacci sequence.
	For any $p\geq2$, it holds that
	\begin{equation} \label{eq:lemma2_1}
		\phantom{\forall n  \in \mathbb{N}_{\ne 0}}
		\hspace{2.5cm}
		2^{\,\, (j-1)/2} \leq S_{j, \, p} \leq 2^{\, j-1}
		\hspace{2.5cm}
		\forall j  \in \mathbb{N}_{\ne 0}
		\,.
	\end{equation}
	Moreover, in the special case $p = 1$, it holds that
	\begin{equation} \label{eq:lemma2_2}
		\phantom{\forall n  \in \mathbb{N}_{\ne 0}}
		\hspace{3.85cm}
		S_{j,\,1} = j 
		\hspace{3.85cm}
		\forall j  \in \mathbb{N}_{\ne 0}
		\,.
	\end{equation}
\end{lemma}
\begin{proof}
Let $p$ be an integer $\geq 2$, 
$\left( f_{j,\,p} \right)_j$ 
a $p$-bonacci sequence 
and 
$S_{j,\,p}$ its $j$-{th} partial sum. 

\smallskip
To establish the \textit{lower} bound in \eqref{eq:lemma2_1}, we consider the Fibonacci case ($p=2$). It is well-known that the $j$-th Fibonacci partial sum satisfies the following identity
\citep[see, e.g.,][]{vorobiev2002fibonacci}:
\begin{equation}
	S_{j,\,2} = \sum_{k=1}^{j} f_{k,\,2} = f_{j+2, \, 2} - 1 \,.
	\label{eq:fibo444}
\end{equation}
Moreover, the Fibonacci sequence $\left( f_{j,\, 2} \right)_j$ admits the following lower bound:
\begin{equation} 
	\phantom{\forall n \geq 1}
	\hspace{3.25cm}
	f_{j,\,2} \geq 2^{~(j-2)/2}
	\hspace{3.25cm}
	\forall j \geq 1
	\label{eq:fibo555}
\end{equation}
which can be verified by induction. Substituting this inequality into \eqref{eq:fibo444} yields 
\begin{equation} \label{eqn:fibo_sums_lb}
	\phantom{\forall j \geq 1}
	\hspace{2cm} 
	S_{j, \, 2} = 
	f_{j+2, \, 2} - 1 \geq
	f_{j+1, \, 2} \geq
	2^{~(j-1)/2}
	\hspace{2cm} 
	\forall j \geq 1 \,.
\end{equation}
Since the Fibonacci sequence grows more slowly than any $p$-bonacci sequence with $p > 2$,
it follows that $S_{j,\,p} \geq S_{j,\,2} ~~ \forall j \geq 1$.
Hence, the lower bound \eqref{eqn:fibo_sums_lb} applies to every $p$-bonacci sequence with $p \geq 2$.

\medskip
To establish the \textit{upper} bound in \eqref{eq:lemma2_1}, we observe that, by the definitions \eqref{eq:p-bonacci} and \eqref{eq:sump-bonacci}, the terms of a $p$-bonacci sequence satisfy 
$f_{j,\,p} \leq S_{j-1,\,p} ~~ \forall j \geq 2$. 
%
By adding $S_{j-1,\,p}$ on both sides, it follows that
\begin{equation}
	\phantom{\forall j \geq 2}
	\hspace{3.25cm}
	S_{j,\,p}
	\leq 
	2 S_{j-1,\,p}
	\hspace{3.25cm}
	\forall j \geq 2
	\,.
\end{equation}
Iterating this inequality yields, for every $j \geq 2$,
\begin{equation} \label{eqn:sums_ub}
	S_{j,\,p} \leq 
	2 S_{j-1,\,p} \leq 
	4 S_{j-2,\,p} \leq 
	\dots \leq 
	2^{~j-1} S_{1,\,p} \,.
\end{equation}
Since $S_{1,\,p} = f_{1,\,p} = 1$, the result follows.

\medskip
Lastly, the result in \eqref{eq:lemma2_2} is immediate, as all terms of the \( p \)-bonacci sequence for \( p = 1 \) are identically equal to 1.

\end{proof}
%

\clearpage
\newpage

\section*{Notation and shorthands}

\quad \bigskip
\begin{center}
\quad \bigskip
\setstretch{1.20}
\begin{tabular} {|l|l|}
	\toprule
	\textbf{Symbol}& \textbf{Description}\\ \midrule
    \hspace{0.5cm}$\mathcal{A}$ & Activation function \\
    \hspace{0.5cm}$A^{(t)}$ & Derivative of the activation function at time $t$ \\
    \hspace{0.5cm}$\vvv{a}{t}$ & Pre-activation state at time $t$ \\
    \hspace{0.5cm}$\mathbf{b}$ & Input-to-hidden bias \\
    \hspace{0.5cm}$\mathbf{c}$ & Hidden-to-output bias \\
    \hspace{0.5cm}$f_{j,p}$ & $j$-th element of a $p$-bonacci sequence \\
    \hspace{0.5cm}$\vvv{h}{t}$ & Hidden state at time $t$ \\
    \hspace{0.5cm}$h$ & Dimension of hidden state \\
    \hspace{0.5cm}$L$ & Loss function (per-sample) \\
    \hspace{0.5cm}$\mathcal{L}$ & Aggregate loss over all training samples \\
    \hspace{0.5cm}$\mathcal{O}(\bm{\cdot})$ & Leading-order (time and space) complexity \\
    \hspace{0.5cm}$p$ & Number of feedback connections \\
    \hspace{0.5cm}$S_{j,p}$ & $j$-th partial sum of a $p$-bonacci sequence \\
    \hspace{0.5cm}$\tau$ & Length of input sequence \\
    \hspace{0.5cm}$U$ & Input-to-hidden kernel \\
    \hspace{0.5cm}$V$ & Hidden-to-output kernel \\
    \hspace{0.5cm}$W_k$ & Autoregressive kernels \\
    \hspace{0.5cm}$w$ & Total number of weights \\
    \hspace{0.5cm}$\vvv{x}{t}$ & Exogenous input vector at time $t$ \\
    \hspace{0.5cm}$x$ & Dimension of exogenous input vector \\
    \hspace{0.5cm}$\vvv{y}{t}$ & Target output at time $t$ \\
    \hspace{0.5cm}$\vvh{y}{t}$ & Output vector at time $t$ \\
    \hspace{0.5cm}$y$ & Dimension of output vector \\
    \hspace{0.5cm}$\boldtheta$ & Vector of input-to-hidden weights \\
    \hspace{0.5cm}$\vartheta$ & Number of input-to-hidden weights \\
    \hspace{0.5cm}$\boldphi$ & Vector of hidden-to-output weights \\
    \hspace{0.5cm}$\varphi$ & Number of hidden-to-output weights \\
	\bottomrule		
\end{tabular}	
\end{center}	

\bigskip 

\begin{center}
	\setstretch{1.10}
	\begin{tabular}{|c|l|}
		\toprule
		\textbf{Acronym}& \textbf{Description}\\ \midrule
		AI & Artificial Intelligence \\
		AAD & Adjoint Automatic Differentiation \\
		ADF & Augmented Dickey Fuller \\
		APL & Average Pinball Loss \\
		ARX & AutoRegressive model with eXogenous inputs \\
		BPTT & BackPropagation Through Times \\
		ESO & Electricity System Operator \\
		FNN & Feedforward Neural Network \\
		GRU & Gated Recurrent Units \\
		IEC & International Electrotechnical Commission \\
		ISO & International Organization for Standardization \\
		ISONE & Independent System Operator New England \\
		LDDN & Layered Digital Dynamic Network \\
		LSTM & Long Short-Term Memory \\
		MAE & Mean Absolute Error \\
		MAPE & Mean Absolute Percentage Error \\
		MSE & Mean Squared Error \\
		NLL & Negative Log-Likelihood \\
		PACF & Partial AutoCorrelation Function \\
		PI & Prediction Interval \\
		RMSE & Root Mean Square Error \\
		RNN & Recurrent Neural Networks \\
		RTRL & Real-Time Recurrent Learning \\
		SE & Standard Error \\
		\bottomrule
	\end{tabular}
\end{center}

\end{document}